\documentclass[preprint,12pt]{elsarticle}

\usepackage{amssymb}
\usepackage{amsmath}
\usepackage{algorithm}
\usepackage{algpseudocode}
\usepackage{booktabs}
\usepackage{multirow}
\usepackage{color}
\usepackage{graphicx}
\usepackage{subcaption}
\usepackage{subfig}

\usepackage{lineno}

\begin{document}

\begin{frontmatter}



\title{Embedding Space Allocation with Angle-Norm Joint Classifiers for Few-Shot Class-Incremental Learning} 

\author[label1,label3]{Dunwei Tu}
\ead{tudunwei@smail.nju.edu.cn}
\author[label1,label3]{Huiyu Yi}
\ead{211300010@smail.nju.edu.cn}
\author[label1,label3]{Tieyi Zhang}
\author[label1,label3]{Ruotong Li}
\author{Furao Shen\corref{cor1}\fnref{label1,label3}}
\ead{frshen@nju.edu.cn}
\author[label4]{Jian Zhao}

\affiliation[label1]{organization={National Key Laboratory for Novel Software Technology},
            country={China}}

            
\affiliation[label3]{organization={School of Artificial Intelligence},
            addressline={Nanjing University},
            city={Nanjing},
            postcode={210023},
            country={China}}
            
\affiliation[label4]{organization={School of Electronic Science and Engineering},
            addressline={Nanjing University},
            city={Nanjing},
            postcode={210023},
            country={China}}
            
\cortext[cor1]{Corresponding author}



\begin{abstract}
Few-shot class-incremental learning (FSCIL) aims to continually learn new classes from only a few samples without forgetting previous ones, requiring intelligent agents to adapt to dynamic environments.
FSCIL combines the characteristics and challenges of class-incremental learning and few-shot learning: (i) Current classes occupy the entire feature space, which is detrimental to learning new classes. (ii) The small number of samples in incremental rounds is insufficient for fully training.
In existing mainstream virtual class methods, to address the challenge (i), they attempt to use virtual classes as placeholders.
However, new classes may not necessarily align with the virtual classes.
For challenge (ii), they replace trainable fully connected layers with Nearest Class Mean (NCM) classifiers based on cosine similarity, but NCM classifiers do not account for sample imbalance issues.
To address these issues in previous methods, we propose the class-center guided embedding Space Allocation with Angle-Norm joint classifiers (SAAN) learning framework, which provides balanced space for all classes and leverages norm differences caused by sample imbalance to enhance classification criteria.
Specifically, for challenge (i), SAAN divides the feature space into multiple subspaces and allocates a dedicated subspace for each session by guiding samples with the pre-set category centers.
For challenge (ii), SAAN establishes a norm distribution for each class and generates angle-norm joint logits.
Experiments demonstrate that SAAN can achieve state-of-the-art performance and it can be directly embedded into other SOTA methods as a plug-in, further enhancing their performance.
\end{abstract}

\journal{Neural Networks}



\begin{keyword}
few-shot learning \sep incremental learning \sep embedding space allocation \sep prototype learning


\end{keyword}

\end{frontmatter}



\section{Introduction}
Deep Neural Network (DNN) methods have excelled in the field of computer vision \cite{deng2009imagenet,he2016deep,simonyan2014very,tan2020efficientdet}.
Leveraging vast datasets and known classification targets, DNNs have demonstrated remarkable capabilities.
However, in real-life scenarios, datasets are often limited in size, so the application of few-shot learning (FSL) is more prevalent \cite{wang2020generalizing,sung2018learning,snell2017prototypical}.
Moreover, in many application contexts, not only is the number of samples limited, but the number of classes continues to grow. Constrained by computational resources and practical constraints, retraining the entire model from scratch with each new class addition is not feasible.
Therefore, the design of efficient and accurate algorithms for achieving Few-Shot Class-Incremental Learning (FSCIL) has attached increasing attention \cite{tao2020few,kalla2022s3c,zhang2021few,zhou2022forward,yang2023neural,lin2024m2sd}.

\begin{figure}[h]
    \centering
    \includegraphics[width=0.8\textwidth,trim=30 20 30 20,clip]{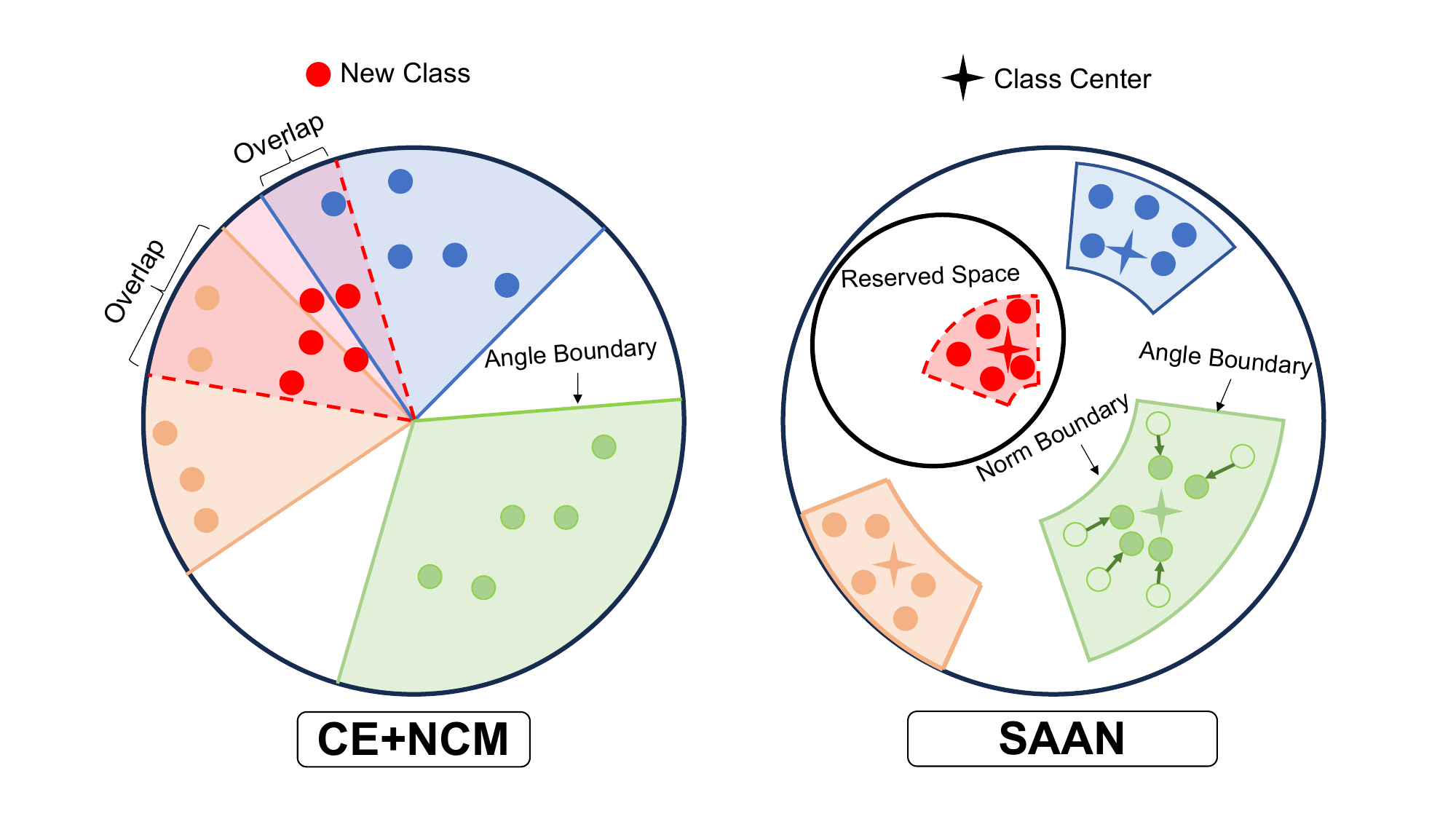}
    \caption{The motivation of SAAN. In only cross-entropy supervision, old categories fill the entire feature space, making it hard for new categories to be inserted without overlapping. In SAAN, the old categories are allocated to a limited space, guided by the category centers, allowing new categories to be inserted into the reserved space without overlapping. In addition, compared to NCM, which has only angular decision boundaries, SAAN has both angular and norm boundaries, which fully utilizes the feature information to enhance the basis for discrimination.}
    \label{fig:motivation}
\end{figure}

The goal of FSCIL is to acquire new knowledge effectively while preventing the forgetting of previously learned knowledge.
Previous research on FSCIL shows that models trained incrementally are prone to overfit on limited new data and forget the old knowledge catastrophically \cite{cheraghian2021semantic,dong2021few,tao2020few}.
To combat overfitting and catastrophic forgetting, some studies propose the use of a feature extractor freezing paradigm \cite{kalla2022s3c,zhang2021few,zhou2022forward,song2023learning}.
This paradigm freezes most layers in the Convolutional Neural Network (CNN), which is the feature extractor, and uses the Nearest Class Mean (NCM) algorithm \cite{mensink2013distance} instead of fully connected layers as the classifier during the incremental phase.
According to the neural collapse phenomenon \cite{kothapalli2022neural,han2021neural}, researchers discover that without any intervention, current training samples would occupy the entire embedding space.
This phenomenon is beneficial for enhancing the neural network's capabilities in conventional classification tasks.
However, in the context of Class-Incremental Learning (CIL), learning new classes becomes extremely challenging.
This phenomenon addresses the challenge (i) we aim to solve.
To address this issue, some researchers \cite{song2023learning,zhou2022forward} propose occupying some of the embedding space with virtual classes generated through data augmentation \cite{yang2024entaugment,yang2024investigating}.
Then, during the incremental phase, the space occupied by these virtual classes can be used for inserting new classes.
However, these methods cannot control the location of the space occupied by virtual classes, as the occupied positions are determined by the features of the virtual samples.
Some researchers \cite{yang2023neural} propose using pre-allocated fixed prototypes to reduce the feature misalignment problem during the incremental process.
However, this approach of completely fixed and randomly allocated prototypes ignores the intrinsic meaning of the samples, which may lead to a discrepancy between the semantic distance of the samples and their distance in the feature space.

Because of the FSL setting, the small number of samples in incremental sessions is insufficient for large-scale training, addressing the challenge (ii) we aim to solve.
For choosing the neural network classifier, NCM \cite{guerriero2018deepncm:} does not rely on extensive training and is suitable for the FSCIL problem, which is wildly used in the mainstream methods \cite{zhang2021few,zhou2022forward,song2023learning}.
NCM solely computes the distance between samples and the mean of each class to classify them, which does not rely on a large amount of data.
Some researchers \cite{thongtan2019sentiment,zhou2022problems} find that cosine similarity is well-suited for comparing embeddings in DNN.
NCM based on cosine similarity has achieved good results in FSCIL \cite{zhang2021few,zhou2022forward,song2023learning}.
In the FSL setting, we believe that leveraging as much sample information as possible during the incremental sessions is essential.
Based on our experiments, we observe that in the FSCIL scenario, there is a significant difference in the number of samples across different categories, leading to variations in norms between categories.
However, this norm information is not used by NCM.

To address these issues, we propose the class-center guided embedding Space Allocation with Angle-Norm joint classifiers (SAAN) learning framework.
SAAN consists of the Class-Center guided embedding Space Allocation (CCSA) and the Angle-Norm Joint classifiers (ANJ). 

CCSA is designed to address the challenge (i) posed by CIL.
Specifically, CCSA divides the feature space into multiple subspaces and allocates different subspaces for different incremental sessions by setting class centers, using our designed cosine center loss.
CCSA resolves two issues found in previous work.
First, in prior virtual class methods \cite{zhou2022forward,song2023learning}, due to the lack of guidance for new classes, the reserved space for virtual classes may not necessarily be utilized by new classes. CCSA imposes explicit spatial constraints on new classes, ensuring that they enter the reserved space.
Second, unlike methods with completely fixed prototype allocation \cite{yang2023neural,ahmed2024orco}, where random allocation may cause similar classes to be far apart in the feature space, SAAN enables the adjustment of class centers within the designated space to ensure that prototype similarity aligns with the samples' semantic information.

ANJ is a new classifier, designed to address the challenge (ii) posed by FSL.
ANJ establishes a norm distribution for each class to estimate the norm logit and combines this with NCM to estimate the angular logit, ultimately generating joint logits. In this way, ANJ can leverage previously overlooked norm information, which is especially important for the information-scarce FSL setting.

Since SAAN is designed to address issues in previous virtual class methods, it can be directly integrated into these State-Of-The-Art (SOTA) methods as a plug-in. {Experiments in both conventional and open-ended scenarios demonstrate that SAAN not only achieves performance comparable to SOTA but also further enhances the performance of existing SOTA methods.}

The main contributions of this paper are summarized as follows:

1) CCSA guides sample points into specific subspaces through our meticulously designed cosine center loss.
This resolves two issues found in previous methods: new classes may fail to appear in the reserved feature space, and similar classes may be far apart in the feature space.

2) Our experiments demonstrate the importance of norm information for classification in the context of FSCIL. ANJ effectively combines angular and norm information, fully utilizing the information contained within the limited samples.

3) Experiments demonstrate that SAAN can achieve performance close to SOTA and further improve SOTA models' performance.
Specifically, SAAN improves the final round accuracy by over 3\% across three datasets and two methods.

\section{Related Works}

\textbf{Few-Shot Learning.} FSL aims to adapt to unseen classes using limited training instances \cite{wang2020generalizing,sung2018learning,snell2017prototypical,fu2021meta}.
The two prominent approaches are optimization-based methods and metric-based methods.
Optimization-based methods \cite{ravi2016optimization,wang2019hybrid,
arnold2021maml,nichol2018first}, situated within meta-learning frameworks, aim to enhance the acquisition of generalizable representations from limited data.
Conversely, metric-based algorithms \cite{oreshkin2018tadam,liu2020negative,zhang2020deepemd,wang2019hybrid,triantafillou2017few,wang2021metric} leverage a pre-trained feature extraction backbone, utilizing distinct distance metrics to ascertain the similarity or dissimilarity between support and query instances.
The separation of tasks into feature extraction and classification has proven effective in FSCIL.

\textbf{Class-Incremental Learning.} CIL is defined as the ongoing incorporation of new knowledge while retaining previously acquired knowledge \cite{mittal2021essentials,masana2022class,zhang2020class}.
Algorithms dedicated to CIL can be broadly divided into four principal categories.
The first approach aims to modify the network's architecture to expand the model's capacity for accommodating new tasks \cite{wang2022learning,yan2021dynamically}.
The second approach focuses on enhancing the distinction between older and newer classes by selectively revisiting old samples \cite{rebuffi2017icarl,zhao2020maintaining,wu2019large}.
The third approach employs knowledge distillation to maintain the model's pre-existing knowledge \cite{rajasegaran2020self,li2017learning,hinton2015distilling}.
The fourth approach evaluates the importance of each parameter to ensure critical ones remain unaltered \cite{aljundi2018memory,liu2018rotate,zenke2017continual}.
Direct application of CIL methodologies to FSCIL is hindered by overfitting issues associated with limited sample sizes.

\textbf{Few-Shot Class-Incremental Learning.}
FSCIL diverges from CIL in its requirement for learning from limited samples for novel classes.
FSCIL is primarily challenged by the catastrophic forgetting of previously learned classes and a propensity for overfitting to new classes.
To enable the model to have forward-compatible learning capabilities, some researchers \cite{zhou2022forward,song2023learning} create virtual classes to reserve space for future classes, while others \cite{yang2023neural,ahmed2024orco} set fixed prototypes for all classes in advance.
FACT \cite{zhou2022forward,ahmed2024orco} proposes a forward-compatible training by assigning virtual prototypes to squeeze the embedding of known classes and reserving for new classes.
Besides, FACT forecasts potential new categories and prepares for the updating process.
The virtual prototypes enable the model to accommodate future updates, serving as proxies dispersed throughout the embedding space to enhance the classifier's performance during inference.
SAVC \cite{song2023learning} innovates by distinguishing new classes from existing ones through the incorporation of virtual classes into supervised contrastive learning.
The virtual classes, generated through predefined transformations, serve not only as placeholders for unseen categories within the representation space but also enrich it with diverse semantic information.
These virtual class methods demonstrate the importance of reserving embedding space for new classes and enhancing the separability of embeddings.

NC-FSCIL \cite{yang2023neural} sets up classifiers with an Equiangular Tight Frame (ETF) structure that considers future classes, and it induces class alignment with the corresponding classifier to mitigate the classifier misalignment issue during the incremental process.
OrCo \cite{ahmed2024orco} maximizes the boundaries between classes and reserves space for subsequent incremental data by introducing perturbed prototypes in the feature space and maintaining orthogonality between classes.
These methods of pre-allocated prototypes highlight the importance of prior space allocation and preventing the misalignment of old categories in new sessions.


{FACT occupies embedding space through virtual prototypes, while SAVC generates negative samples for contrastive supervised learning via data augmentation to preserve embedding space. Additionally, contrastive learning enhances the model's ability to generalize, which is crucial in FSL. NC-FSCIL assigns fixed prototypes to categories randomly before training and uses prototype alignment as the supervisory signal for the model. SAAN is a more flexible and semantically aligned learning strategy based on similarity estimation for spatial pre-allocation and dynamic adjustment of prototypes.}



\section{Preparatory Work}
\subsection{Problem Setting}
In FSCIL, the model is fed with a sequence of training dataset $\left\{ \mathcal{D}_{train}^{t}\right\}_{t=0}^{T}$, where $\mathcal{D}_{train}^{t}$ is the training set at the session $t$ and $T$ is the number of all sessions.
$\mathcal{D}_{train}^{t}=\left\{ (\mathbf{x}_i,y_i)\right\}$, where $\mathbf{x}_i$ is the $i$-th training sample and $y_i$ is the $i$-th label.
$\left|\mathcal{D}_{train}^t\right|$ denotes the number of the training samples at session $t$.
Each session $t$ has a distinct label space $\mathcal{Y}^t$, which means that $\mathcal{Y}^{t} \cap \mathcal{Y}^{t^{\prime}}=\varnothing$ for any $t \neq t^{\prime}$.
$\left|\mathcal{Y}^t\right|$ denotes the number of the classes at session $t$.
In the base session, $\left|\mathcal{D}_{train}^0\right|$ is equal to the number of all the samples whose labels $\in \mathcal{Y}^{0}$ in the dataset.
For example, on CIFAR100 \cite{krizhevsky2009learning}, there are 500 training samples for each class at the base session.
However, at the incremental sessions, we only can access a limited number of training samples.
The training data is organized in an $N$-way $K$-shot format which means there are $N$ classes, and each class contains $K$ training samples at each incremental session.
For example, on CIFAR100, each incremental session includes 5 classes, with each class having 5 samples, meaning $N=5$ and $K=5$.
At each session $t$, the model is tested on all test samples whose labels $\in (\mathcal{Y}^{0} \cup \mathcal{Y}^{1} \cup \dots \cup \mathcal{Y}^{t})$.

\subsection{The Model Framework}
The whole model can be decoupled as a feature extractor and a classifier which follows \cite{tao2020few,kalla2022s3c,zhang2021few,zhou2022forward,song2023learning}.
We employ a CNN as the feature extractor, denoted by $f(\cdot)$.
The output of this CNN, $\mathbf{e}_i = f(\mathbf{x}_i)$, serves as the basis for classification.
In the feature extractor frozen paradigm, the CNN is fully trained at the base session and fine-tuned on the basis of freezing most of the layers in CNN at the incremental sessions.
The NCM classifier recognizes samples by: \begin{equation}
\label{eq:NCM_classify}
 y_\mathbf{x}^\text{p}=\arg \max _{j} \ \cos\langle \mathbf{e},\boldsymbol{w}_{j}^{t}\rangle,
\end{equation} 
where $y_{\mathbf{x}}^{\text{p}}$ is the predicted class, $\cos\langle\cdot, \cdot\rangle$ is cosine similarity function, and $\boldsymbol{w}_{j}^{t}$ is the representative point of the class $j$ at the session $t$.

The representative point $\boldsymbol{w}_{j}^{t}$ in NCM classifier is calculated by: \begin{equation}
\label{eq:NCM_point}
 \boldsymbol{w}^t_j = \frac{1}{n^t_j}\sum^{n^t_j}_{i=1} \left(\mathbf{e}_{i}\mathbb{I}(y_i=j)\right),
\end{equation}  where ${n^t_j}$ is the number of samples of class $j$ and $\mathbb{I}\left(\cdot\right)$ is the indicator function.
As indicated by Eq. \ref{eq:NCM_classify}, the NCM classifier, which classifies based on the nearest neighbor principle by measuring the cosine similarity between sample embeddings and representative points, completely disregards the norm information of the sample embeddings.

\renewcommand{\Comment}[1]{\hfill// #1}
\begin{algorithm}
\caption{{The training and classification process for the feature-freezing base approach.}}
\begin{algorithmic}[1]

\Require The CNN $f(\cdot)$, The classifier $W$, The training set sequence $\left\{ \mathcal{D}_{train}^{t}\right\}_{t=0}^{T}$, The model learning rate $\mu$
\Ensure The updated CNN $f(\cdot)$, The updated classifier $W$

\For{each training dataset $\mathcal{D}_{train}^{t}$}
    \State Expand the dimension of $W$ from $d \times |\mathcal{Y}^{0} \cup \mathcal{Y}^{1} \cup \dots \cup \mathcal{Y}^{t-1}|$ to $d \times |\mathcal{Y}^{0} \cup \mathcal{Y}^{1} \cup \dots \cup \mathcal{Y}^{t}|$
    
    \If{$t>0$}
        \State Freeze most layers of the CNN for fine-tuning preparation
    \EndIf
    
    \While{$\mathcal{L}_{ce}$ not converges}
        \For{each batch \{$\mathbf{x}_1, \mathbf{x}_2,\dots,\mathbf{x}_m\} \in \mathcal{D}_{train}^{t}$}
            \State $e \leftarrow e+1$
            \State Compute $\mathcal{L}_{ce}$ on $\{\mathbf{x}_1, \mathbf{x}_2,\dots,\mathbf{x}_m\}$
            \State$\theta_f^{e+1} \leftarrow\theta_f^e- \mu\frac{\partial \mathcal{L}_{ce}}{\partial \theta_f^e}$ 
            \State$W^{e+1} \leftarrow W^e- \mu\frac{\partial \mathcal{L}_{ce}}{\partial W^e}$ 
            
            
        \EndFor
    \EndWhile
    \State Recalculate $W$ by Eq. \ref{eq:NCM_point} to replace fully connected layers by NCM
    \State Classify all the seen samples using the continuously expanding $W$ and the continuously updating $f$ through Eq. \ref{eq:NCM_classify}
\EndFor
\State \Return $f(\cdot)$, $W$
\end{algorithmic}
\label{alg:feature_freezing}
\end{algorithm}
\subsection{Learning of the Baseline Model}
The optimization goal of the feature extractor freezing paradigm baseline is to minimize \(\mathcal{L}_{ce}(\phi(\mathbf{x}), y)\), where \(\phi(\mathbf{x}) = W^\top f(\mathbf{x})\) is the model composed of the feature extractor and the linear classifier, and \(\mathcal{L}_{ce}(\cdot)\) is the cross-entropy loss function.
Here, $\phi(\mathbf{x})\in \mathbb{R}^{|\mathcal{Y}|\times 1}$, $W \in \mathbb{R}^{d\times |\mathcal{Y}|}$, $f(\mathbf{x})\in \mathbb{R}^{d \times 1}$, and $d$ is the feature dimension.
After the base session, most layers of the CNN are frozen and fine-tuned.
The NCM classifier expands by $W=\{\boldsymbol{w}^{0}_{1},\boldsymbol{w}^{0}_{2},\cdots,\boldsymbol{w}^{t}_{|\mathcal{Y}^{t}|-1},\boldsymbol{w}^{t}_{|\mathcal{Y}^t|}\}$.
\(\boldsymbol{w}^{t}\) is continuously expanded during the incremental process based on new data.
Although the feature-freezing base approach is simple, it outperforms many FSCIL methods, making it a strong baseline model.
{The training and classification algorithm depicting the feature-freezing base approach is shown in Alg. \ref{alg:feature_freezing}.}

\begin{figure*}
  \centering
  \includegraphics[width=\linewidth, trim=0 40 20 70, clip]{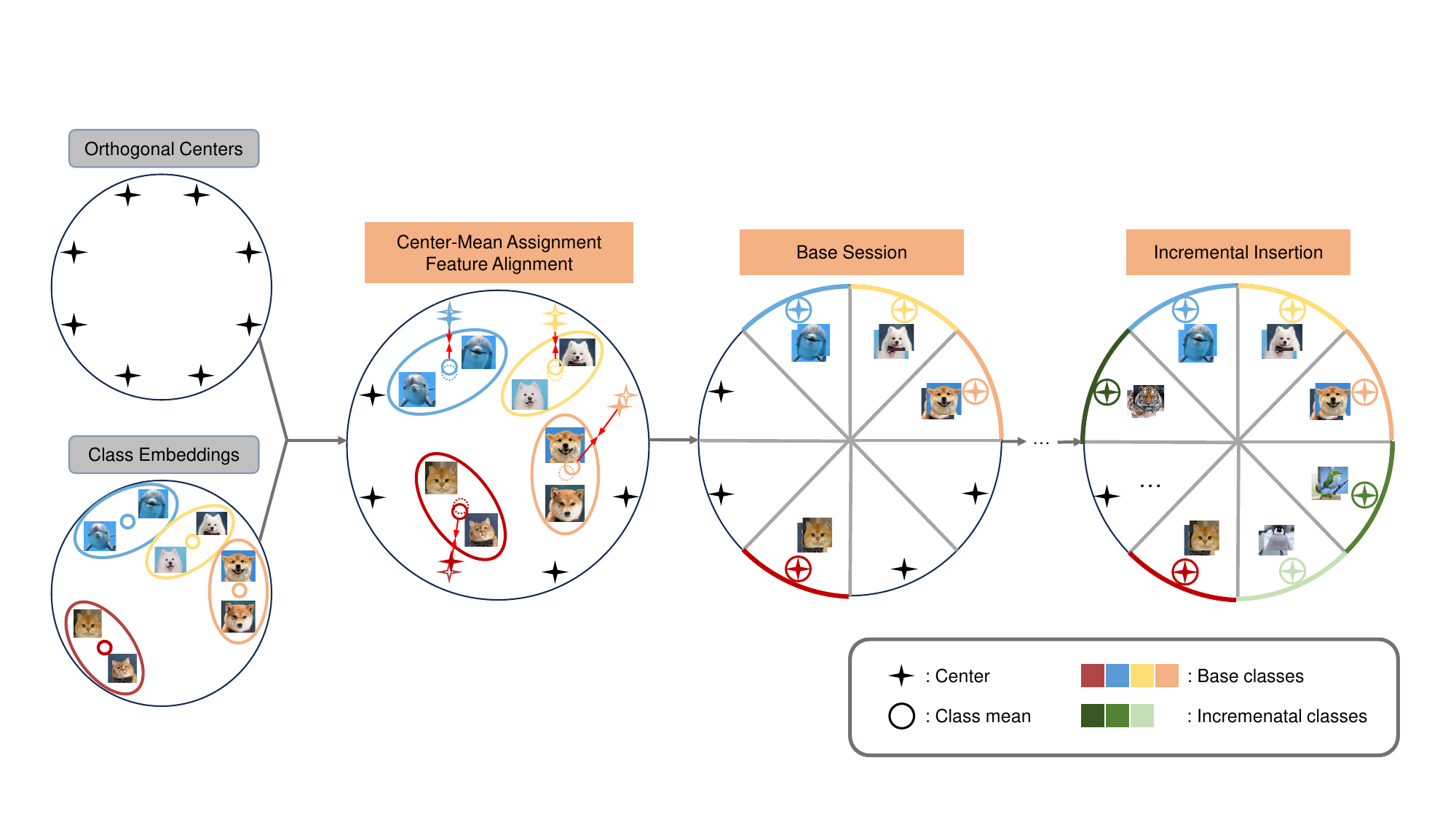}
  \caption{
  Illustration of CCSA. CCSA sets orthogonal class centers and assigns these centers to different categories to allocate the embedding space. CCSA uses the well-designed cosine center loss to guide embeddings into the corresponding subspace, achieving the effect of feature alignment. The centers are adjusted based on semantics in the base session, and newly inserted classes are allocated to the remaining space during the incremental sessions.}
  
  \label{fig:CCSA}
\end{figure*}

\section{Method}

The whole method, the class-center guided embedding Space Allocation with Angle-Norm joint classifiers (SAAN), consists of the Class-Center guided embedding Space Allocation (CCSA) and the Angle-Norm Joint classifiers (ANJ), which are illustrated in Fig. \ref{fig:CCSA} and Fig. \ref{fig:ANJ}.
CCSA and ANJ address the problems of guiding feature learning and classifier construction in FSCIL, respectively.

\subsection{Class-Center Guided Embedding Space Allocation}
The baseline model trained solely with cross-entropy loss focuses only on the current sample during training and utilizes the entire embedding space, rendering it challenging for the model to assimilate new class knowledge \cite{kothapalli2022neural,han2021neural}.
The embedding space consists of all possible feature vectors. This space not only needs to accommodate the feature vectors of the current samples but also needs to reserve capacity for future classes.
Inspired by the center loss \cite{wen2016discriminative}, we mitigate this challenge by establishing learnable class centers (prototypes) and guiding sample embeddings toward their corresponding class centers.
Unlike center loss, CCSA constrains the allocation of center positions and the update of centers to achieve the goal of constraining the model's utilization of the embedding space and providing balanced space for new classes. The overall process of CCSA is shown in Fig. \ref{fig:CCSA}.

\subsubsection{Space allocation}
We first set \(( |\mathcal{Y}^{0} \cup \mathcal{Y}^{1} \cup \dots \cup \mathcal{Y}^{t}|) \) orthogonal class centers and assign them one-to-one based on their distance from the class means.
{In the real world, considering the total number of classes is unknown and continuously growing, we just set $d$ orthogonal centers, which is the maximum number of orthogonal centers that can be generated in a $d$-dimensional feature space.}
To ensure that the total distance between all class centers and their corresponding class means is minimized, we use the Hungarian algorithm \cite{kuhn1955hungarian,munkres1957algorithms} to assign the class centers to different categories.
The Hungarian algorithm is used to solve the assignment problem in a bipartite graph, aiming to find an optimal matching that minimizes the total cost.
{Here, the cost is defined by cosine distance, because the fully connected layer in a neural network performs dot product operations, and some literature \cite{goldberg2014word2vec,mikolov2013distributed} suggests that dot products can introduce bias due to different sample sizes. Cosine distance, on the other hand, can eliminate this bias by removing the influence of vector norms.}
We define an \( ( |\mathcal{Y}^{0} \cup \mathcal{Y}^{1} \cup \dots \cup \mathcal{Y}^{t}|) \times ( |\mathcal{Y}^{0} \cup \mathcal{Y}^{1} \cup \dots \cup \mathcal{Y}^{t}| ) \) cost matrix \( D = [c_{i,j}] \), where \( c_{i,j} \) represents the cosine distance from  \( i \)-th class center to the \( j \)-th class.
Our objective is to find a one-to-one matching that minimizes the total cost.
Mathematically, this can be written as:
\[
\min_{\sigma} \sum_{i=1}^{n} c_{i, \sigma(i)},
\]
where \( \sigma \) is a permutation of the set \( \{1, 2, \ldots,  ( |\mathcal{Y}^{0} \cup \mathcal{Y}^{1} \cup \dots \cup \mathcal{Y}^{t}| )\} \), ensuring that each class center is uniquely assigned to a class and each class is uniquely assigned to a class center.
Since the number of class centers in the base round exceeds the number of existing categories, we set up some virtual classes and assign a distance of 0 between these virtual class means and any class center to prioritize the allocation of existing categories.
After the generation and assignment of centers, sample features are also distributed into different spaces, ultimately achieving the goal of spatial allocation.

\subsubsection{Feature alignment}
We design a novel cosine center loss to guide sample embeddings closer to their corresponding class centers and away from other class centers in the angular space, achieving the effect of feature alignment.
We define the cosine center loss $\mathcal{L}_\text{cc}$ as $\alpha\mathcal{L}_{1}+\beta\mathcal{L}_{2},$ where $\mathcal{L}_1$ pulls embeddings closer to their corresponding centers, $\mathcal{L}_2$ pushes embeddings away from other class centers, and $\alpha$ and $\beta$ are weighting coefficients.
$\mathcal{L}_1$ is defined by:
\begin{equation}
\mathcal{L}_{1}=\frac{1}{m}\sum_{i=1}^{m}\left(1-\cos\langle \mathbf{e}_i,\mathbf{c}_{y_i}\rangle \right),
\end{equation}
where $m$ is the batch size, and $\mathbf{c}_{y_i}$ is the class center of class $y_i$.
$\mathcal{L}_2$ is defined by:
\begin{equation}
\mathcal{L}_{2}=\frac{1}{m}\frac{1}{|\mathcal{Y}|}\sum_{i=1}^{m}\sum_{j=1}^{|\mathcal{Y}|}\left(\cos\langle \mathbf{e}_i,\mathbf{c}_{j}\rangle\mathbb{I}\left( y_i\neq j\right)\right).
\end{equation}
$\mathcal{L}_1$ optimizes the cosine distance between sample embeddings and their corresponding class centers, restricting embeddings near the class centers to achieve the goal of space allocation.
$\mathcal{L}_2$ increases the cosine distance between sample embeddings and centers of other classes.
In addition to space allocation, $\mathcal{L}_\text{cc}$ also plays a role in reducing intra-class distance and improving inter-class separation.
The detailed gradient analysis is provided in Sec. \ref{sec:Gradient_Analysis}.
Under the guidance of class centers, the sample features will enter specific spaces, while retaining some space for the insertion of new classes.

\subsubsection{Center updates}
The pre-assigned and fixed centers may not semantically match the classes.
In other words, semantically close centers may be far apart in embedding space.
To ensure the semantic coherence of the centers, we combine CE loss and cosine center loss to jointly guide the model and adjust the centers during the learning process.
The total loss is defined as:\begin{equation}
    \mathcal{L}_{total}=\mathcal{L}_{ce}(\phi(\mathbf{x}), y)+\alpha \mathcal{L}_{1}+\beta\mathcal{L}_{2}.
\end{equation}
To prevent unreasonable fixed center settings, we apply momentum adjustments to the centers by: \begin{equation}
    \mathbf{c}^{e+1}_j=\mathbf{c}^e_j + \eta\Delta \mathbf{c}_j,
    \label{eq:center_update}
\end{equation}
where $\eta$ is the center adjustment speed, and $\Delta \mathbf{c}_j$ is the momentum update.
$\Delta \mathbf{c}_j$ is obtained by:\begin{equation}
    \arg\max_{\Delta\mathbf{c}_j} \sum_{i=1}^{m}\cos\langle \mathbf{e}_i,\Delta\mathbf{c}_j\rangle\mathbb{I}(y_i=j).
\end{equation}
$\Delta \mathbf{c}_j$ is the point closest to all samples of the same class in terms of cosine distance.
We take a simple solution of $\Delta\mathbf{c}_j$ as:\begin{equation}
    \Delta\mathbf{c}_j = \frac{1}{m}{\sum_{i=1}^{m}\left( \hat{\mathbf{e}}_i\mathbb{I}(y_i=j)\right)},
\end{equation}
where $\hat{\mathbf{e}}_{i}=\frac{\mathbf{e}_{i}}{\|\mathbf{e}_{i}\|}$.
Continuously adding the mean of normalized embeddings to the center will gradually move the center closer to the sample embeddings which is learned by cross-entropy loss.
Setting the centers as learnable parameters and updating them through gradient descent is equivalent to using Eq. \ref{eq:center_update}, as shown in \ref{sec:Gradient_Analysis}.

Allowing the centers to be updated freely, as in center loss, would compromise the spatial allocation function.
Therefore, we let \(\eta\) decay rapidly after each epoch to ensure limited adjustment of the centers.

\subsubsection{Incremental sessions insertion}
In the incremental sessions, we continue to use the Hungarian algorithm to assign the class means of the current session to the unused class centers, and we reset \(\eta\) so that the centers can move within limited bounds.
It should be noted that the already assigned class centers and old class means need to be removed, and the virtual class centers for future classes are retained.
Due to the scarcity of incremental samples, only $\mathcal{L}_1$, which restricts the space, is activated, while $\mathcal{L}_2$, which improves separation, is deactivated.

\begin{figure*}
  \centering
  \includegraphics[width=\linewidth, trim=50 48 30 110, clip]{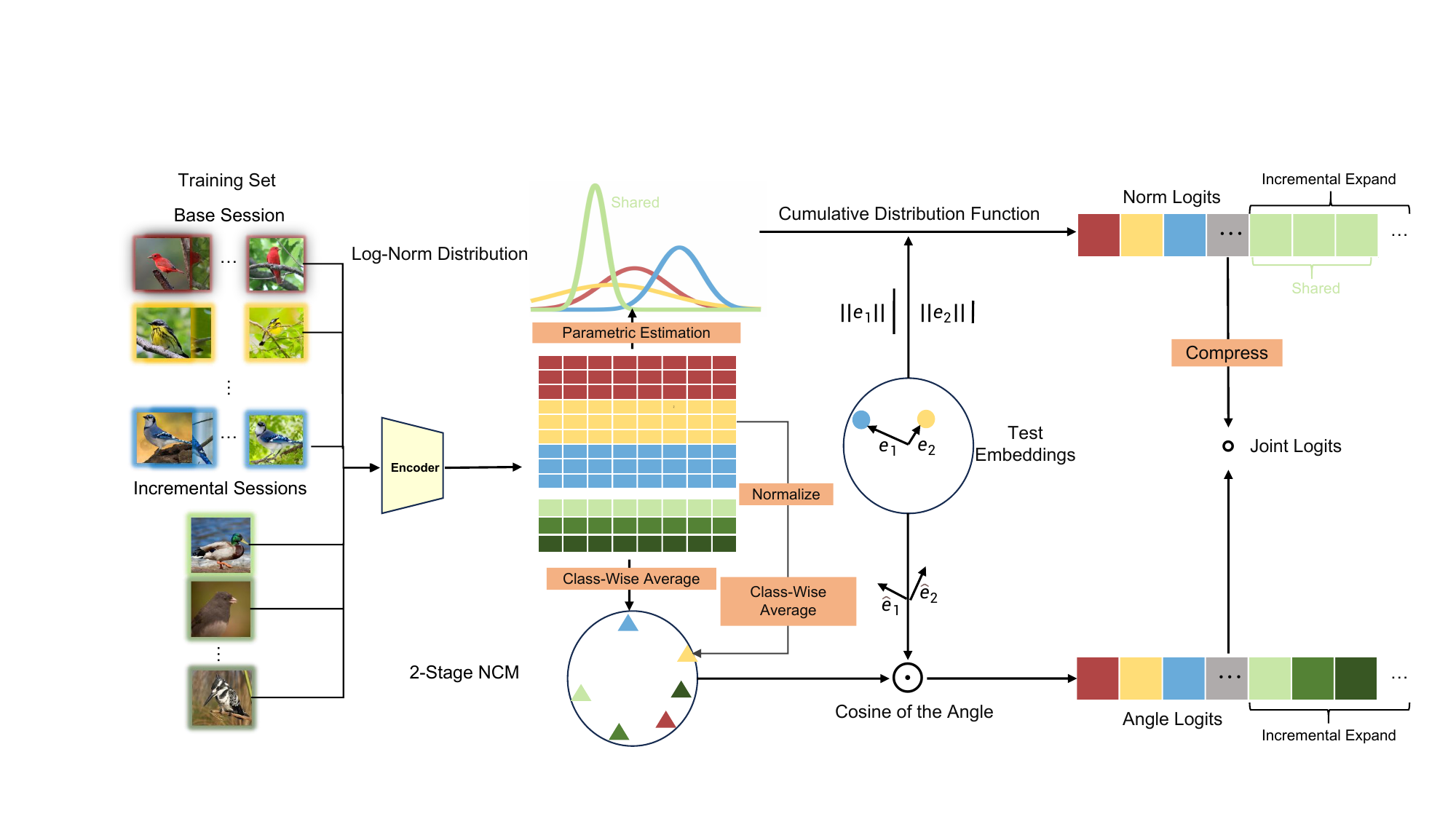}
  \caption{Overall pipeline of ANJ. ANJ consists 2SNCM and Norm Distribution. 2SNCM is fairer than NCM in determining representative points, while Norm Distribution utilizes norm information to establish a probability model, enhancing the basis for classification.}
  \label{fig:ANJ}
\end{figure*}

\subsection{Angle-Norm Joint Classifiers}
We aim to explore the knowledge embedded in the norms, which is ignored by NCM, by establishing a more comprehensive classifier that leverages both angular and norm information simultaneously.
\subsubsection{Log-Norm distribution of embeddings}

\begin{figure}
  \centering
  \begin{subfigure}{0.48\textwidth}
    \includegraphics[width=\linewidth, clip]{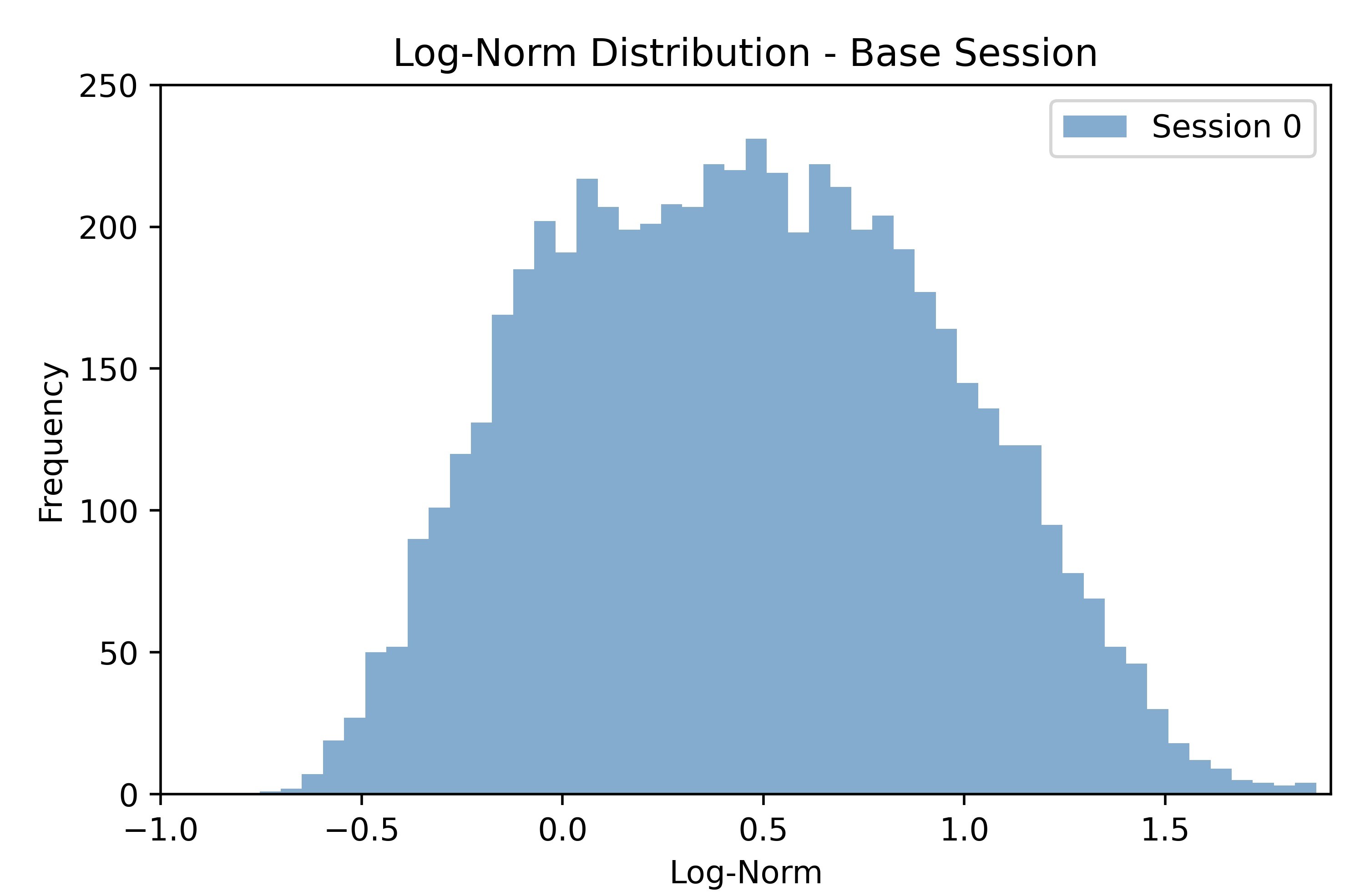}
    \caption{}
    \label{fig:norm_a}
  \end{subfigure}
  \begin{subfigure}{0.48\textwidth}
    \includegraphics[width=\linewidth, clip]{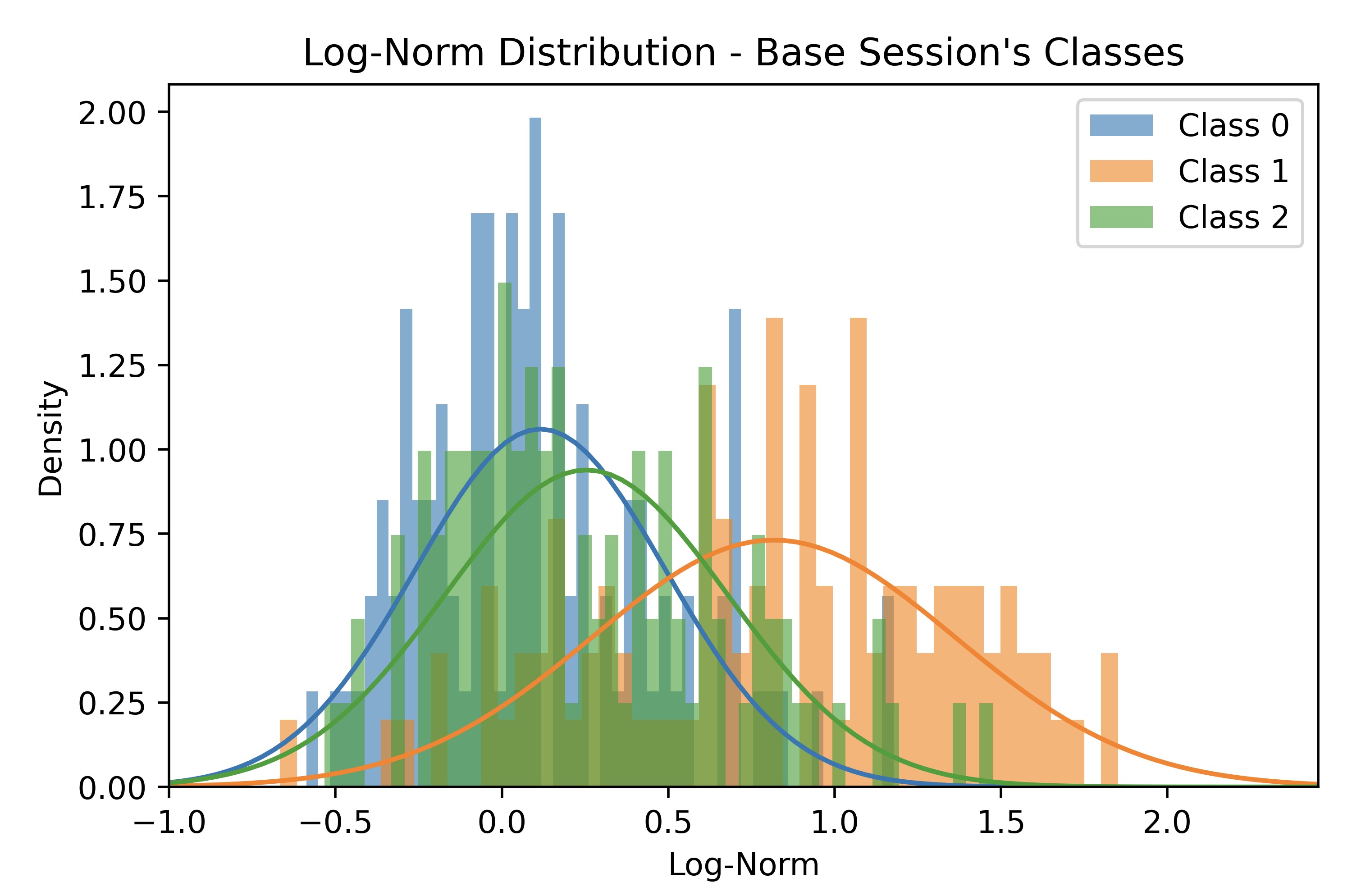}
    \caption{}
    \label{fig:norm_b}
  \end{subfigure}
    \begin{subfigure}{0.48\textwidth}
    \includegraphics[width=\linewidth, clip]{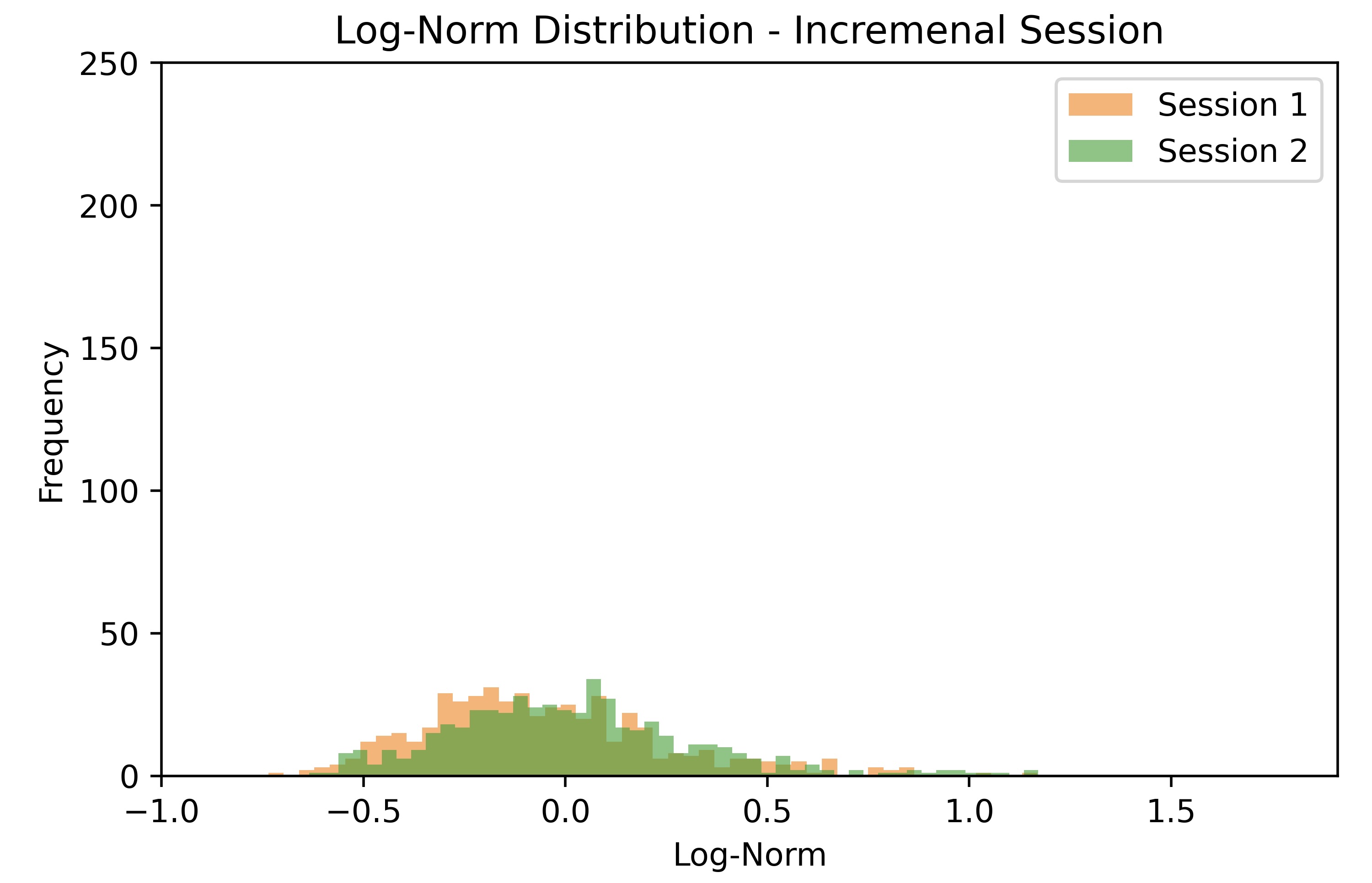}
    \caption{}
    \label{fig:norm_c}
  \end{subfigure}
  \begin{subfigure}{0.48\textwidth}
    \includegraphics[width=\linewidth,clip]{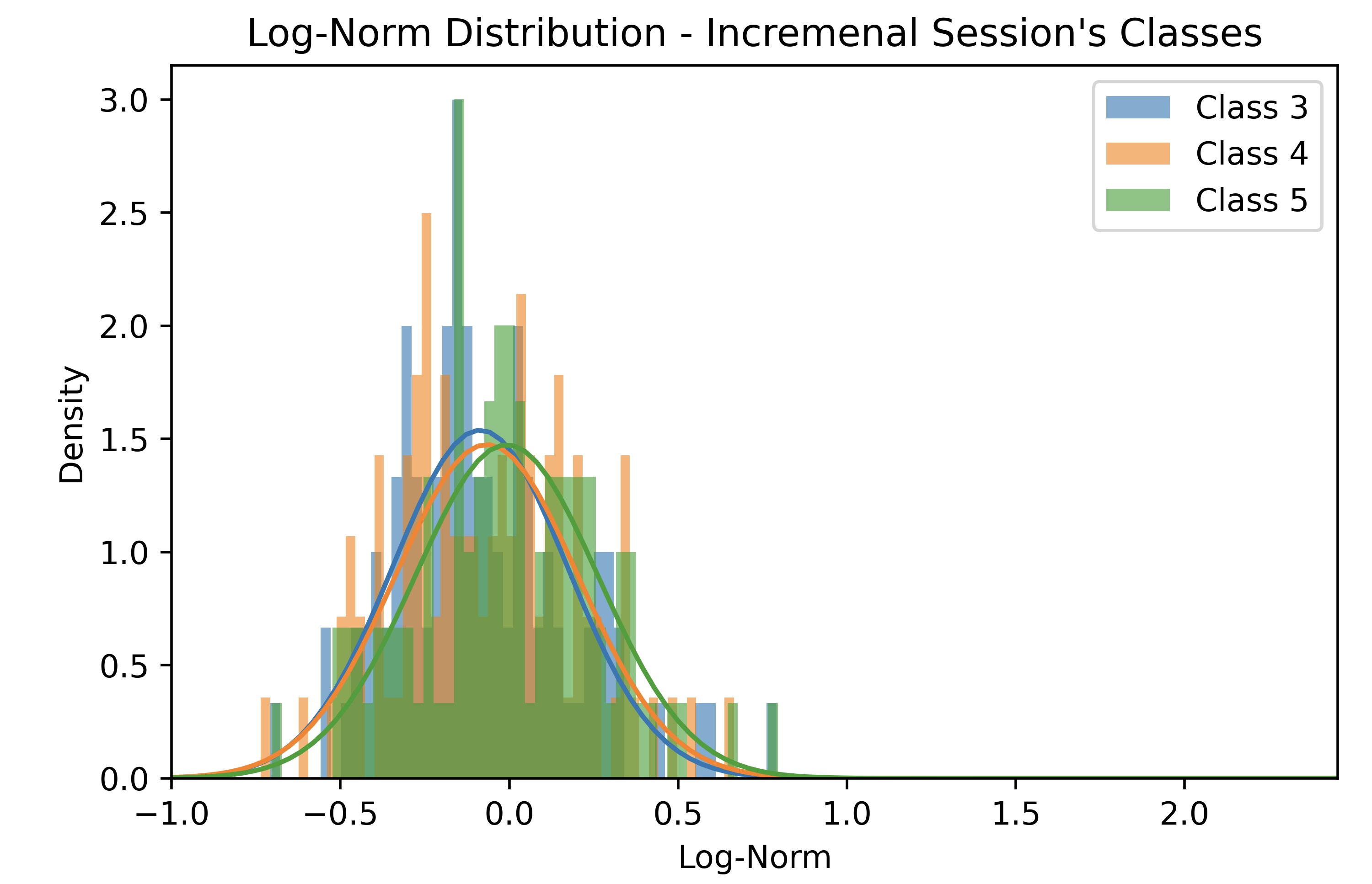}
    \caption{}
    \label{fig:norm_d}
  \end{subfigure}
  \caption{{Histogram of the distribution of the log norms of sample embedding on \emph{mini}ImageNet dataset. (a) The base session. (b) 2 incremental sessions. (c) 3 randomly selected classes in the base session. (d) 3 randomly selected classes in incremental sessions. The curve represents the normal distribution formed by the sample mean and variance of the log norms.}}
  \label{fig:distribution}
\end{figure}

NCM is based on cosine similarity, so it is an algorithm that relies on the angular relationship between vectors while ignoring the vector norm (the distance between the vector's tip and the origin).
We attempt to incorporate this vector length information into the classification criteria.
If the norms of samples from different categories have distinct distributions, it indicates that norms contain label information to a certain extent.
The distributions of $\ln\|\mathbf{e}\|$ from different sessions and categories, as extracted by a CNN trained in the FSCIL paradigm, are shown in Fig. \ref{fig:distribution}.
{In Fig. \ref{fig:norm_a} and \ref{fig:norm_c}, it can be observed that the mean and variance of embedding log norms during the base sessions are significantly higher than those during the incremental sessions.
Based on the characteristic demonstrated by the embedding log norm, we attempt to leverage this feature to assist us in determining whether this class belongs to the base session or the incremental sessions during inference.
Additionally, as shown in Fig. \ref{fig:norm_b} and \ref{fig:norm_d}, in the base session, the log-norm distributions of different classes vary significantly, whereas in the incremental sessions, this difference becomes much smaller. We believe that the variation in log-norm distributions may be caused by the model’s varying degrees of fit to different categories.}
Although there is significant overlap in the norms across different categories and sessions, preventing direct prediction based solely on norms, norms can still serve as a part of the basis for classification.

\subsubsection{Dual-Granularity parametric estimation}

As shown in Fig. \ref{fig:distribution}, we observe significant differences between the incremental and base sessions, but the distributions across different incremental sessions are approximately similar. Based on this phenomenon, we choose to estimate a distribution for the norms of each class.
{Based on the trend of class distributions in Fig. \ref{fig:norm_b} and \ref{fig:norm_d}}, we assume that the log norms of the embeddings for samples belonging to the same category follow a normal distribution, represented as $\ln{\|\mathbf{e}_i\|} \sim \mathcal{N}(\mu_j, \sigma^2_j)$, where $y_i = j$, $j$ is the category label, $\mu_j$ denotes the mean, and $\sigma^2_j$ denotes the variance.

During the base session, with sufficient training samples, we approximate the population mean and variance with sample mean and variance, respectively.
As shown in Fig. \ref{fig:norm_b}, the differences in log norms between classes are significant, making it necessary to perform class-level modeling of the log norms.
During the incremental sessions, however, each class contains only a few samples, which may result in significant deviations between the sample mean and variance and the true population values.
Despite this, as shown in Fig. \ref{fig:norm_d}, the differences in log norm between classes during these sessions are actually quite small.
{We assume that the log norm of each incremental category follows the same normal distribution. We know that if the distribution of samples in each dataset follows the same normal distribution, then mixing the datasets will result in the samples still following a normal distribution,} denoted as $\ln{\|\mathbf{e}_i\|} \sim \mathcal{N}(\mu_\text{incre}, \sigma^2_\text{incre})$, where $y_i\in (\mathcal{Y}^{1} \cup \dots \cup \mathcal{Y}^{t})$, $t\geq 1$, and $\mu_\text{incre}, \sigma^2_\text{incre}$ are the shared mean and variance, respectively.
{We conduct a test in Sec. \ref{sec:normal_test} to determine whether the classes follow a normal distribution.}
We refer to assigning the same distribution to different classes in the incremental sessions as the shared distribution technique.
{A session-granular distribution does not directly improve classification accuracy. Its primary purpose is to distinguish between the base session and incremental sessions. By improving the accuracy of session identification, it indirectly enhances the final classification accuracy.}
In summary, we model the distributions of the base and incremental rounds with different levels of granularity.

\subsubsection{Angle-Norm joint logits inference}
The pipeline of Angle-Norm joint (ANJ) inference is illustrated in Fig. \ref{fig:ANJ}.
ANJ consists of 2-Stage NCM (2SNCM) and Norm Distribution.

\textbf{2-Stage NCM.}
We propose a variant of NCM, named 2-Stage NCM (2SNCM), to utilize angular information.
For 2SNCM, the logits for the $j$-th class $\mathbf{z}_j^1$ are given by $\mathbf{z}_j^1=\cos\langle\mathbf{e},\boldsymbol{w}_j\rangle$, which is consistent with NCM.
The $\boldsymbol{w}_j$ is stage-dependent and is calculated by:\begin{equation}
    \boldsymbol{w}_j^t = \begin{cases} 
        \frac{1}{n^t_j}\sum^{n^t_j}_{i=1} \left(\hat{\mathbf{e}}_{i}\mathbb{I}(y_i=j)\right) & \text{if } t = 0 \\
        \frac{1}{n^t_j}\sum^{n^t_j}_{i=1} \left(\mathbf{e}_{i}\mathbb{I}(y_i=j)\right) & \text{if } t \geq 1
        \end{cases},
\end{equation}
where $\hat{\mathbf{e}}_{i}=\frac{\mathbf{e}_{i}}{\|\mathbf{e}_{i}\|}$.
During the base session, we normalize embeddings to treat each sample equally, because otherwise, samples with larger norms would have greater weight in the calculation of $\boldsymbol{w}_j$.
As illustrated in Fig. \ref{fig:norm_a}, there is considerable variance in the norms of samples during the base session.
In the incremental sessions, due to the scarcity of samples, normalizing embeddings could lead to a significant loss of information.
Therefore, we adopt a staged approach to better accommodate FSCIL.

\textbf{Norm Distribution.}
Based on the distribution of class norms and the norms of sample embeddings, we can obtain the norm logits of $j$-th class for $i$-th sample $\mathbf{z}_j^2$ by:\begin{equation}
    \mathbf{z}_j^2 =\begin{cases}  p\left(X \geq \ln\|\mathbf{e}_i\| \mid X \sim \mathcal{N}(\mu_j, \sigma^2_j)\right)  & \text{if } \ln\|\mathbf{e}_i\| \geq \mu_j \\
    p\left(X \leq \ln\|\mathbf{e}_i\| \mid X \sim \mathcal{N}(\mu_j, \sigma^2_j)\right) & \text{if } \ln\|\mathbf{e}_i\| < \mu_j
\end{cases},
\end{equation}
where $X$ represents a random variable that follows the distribution of norms for class $j$.
We can obtain the probability by the cumulative distribution function of normal distribution.
{We also explore other distributions in Sec. \ref{sec:Ablation}.}

\textbf{Angle-Norm Joint Inference.}
Since norms contain less information than angles, we first compress the norm logits, ultimately obtaining the joint logits.
The final angle-norm joint logits $\mathbf{z}_j$ is calculated by $\mathbf{z}_j = \mathbf{z}_j^1 \left(\mathbf{z}_j^2\right)^C$, where $C$ is the compression coefficient.
After compression, the differences between the norm logits of different categories are reduced, thereby amplifying the role of the angle logits.
The final predicted category is determined by $\mathbf{z}_j$.

\section{Experiments}
\subsection{Experimental Setup}
\label{sec:Experimental Setup}
\textbf{Datasets.} {Following the previous works setting \cite{tao2020few, zhou2022forward,kalla2022s3c, song2023learning}, we perform conventional experiments on three datasets: CIFAR100 \cite{krizhevsky2009learning}, CUB200 \citep{wah2011caltech} and \emph{mini}ImageNet \cite{russakovsky2015imagenet}. To better reflect real-world situations, we also conduct open-ended and imbalanced experiments on the CompCars \cite{yang2015large} dataset.}
CIFAR100 consists of 100 different classes, with each class containing 600 images.
CUB200 contains images of 200 bird species.
\emph{mini}ImageNet is a subset of ImageNet \cite{deng2009imagenet} and consists of 100 selected classes.
{CompCars is a fine-grained vehicle dataset that contains 431 categories, and the large number of categories allows us to set up long-term learning experiments to simulate open-ended scenarios.}
In CIFAR100 and \emph{mini}ImageNet, the base session consists of 60 classes, while the incremental sessions comprise 40 classes.
During the incremental sessions, the data is organized in a 5-way 5-shot format.
As for CUB200, the base session includes 100 classes, and the incremental sessions involve 100 classes as well.
In the incremental sessions of CUB200, the data is organized in a 10-way 5-shot format.

\textbf{Implementation Details.}
We conduct experiments by separately employing SAAN, and embed it into SOTA methods, $i.e.$ FACT \cite{zhou2022forward}, SAVC \cite{song2023learning}.
Both of these methods adopt the feature freezing paradigm and use NCM classifier.
{For fairness, all these methods adopt ResNet18 \cite{he2016deep} for \emph{mini}ImageNet, CUB200 and CompCars, and ResNet20 \cite{he2016deep} for CIFAR100.
In SAAN, the initial learning rate is set to 0.1 with cosine annealing on CIFAR100,\emph{mini}ImageNet, and milestone decay on CUB200 and CompCars.}
The batch size is 256, and training is conducted for 600 epochs.
The training parameters in the embedding of FACT and SAVC are set to be the same as those in their works.
The selection of hyper-parameters used in SAAN is discussed in Sec. \ref{sec:parameter_sensitivity}.

\begin{table}
\centering
\caption{Performance comparison on CUB200 dataset. *: FSCIL results implemented by \cite{tao2020few}. $\ddagger$:reproduced under the FSCIL setting with the source code. Drop: The drop in Acc. of the last session relative to session 0.}
\setlength{\tabcolsep}{1mm}
\resizebox{\textwidth}{!}{%
\begin{tabular}{ccccccccccccc}
\toprule
\multirow{2}{*}{Method} & \multicolumn{11}{c}{Acc. in each session (\%)$\uparrow$} & \multirow{2}{*}{Drop$\downarrow$}\\
\cline{2-12} & 0 & 1 & 2 & 3 & 4 & 5 & 6 & 7 & 8 & 9 & 10 \\
\hline
Finetune* & 68.68 & 43.70 & 25.05 & 17.72 & 18.08 & 16.95 & 15.10 & 10.06 & 8.93 & 8.93 & 8.47 & 60.21  \\
iCaRL* \cite{rebuffi2017icarl} & 68.68 &52.65 &48.61 &44.16 &36.62 &29.52 &27.83 &26.26 &24.01 &23.89 &21.16 &47.52\\
EEIL* \cite{castro2018end} &68.68 &53.26 &47.91 &44.20 &36.30 &27.46 &25.93 &24.70 &23.95 &24.13 &22.11 &46.57\\
TOPIC \cite{tao2020few} &  68.68 & 62.49 & 54.81 & 49.99 & 45.25 & 41.40 & 38.35 & 35.36 & 32.22 & 28.31 & 26.28 &42.40 \\
Decoupled-Cosine$^{\ddagger}$ \cite{NIPS2016_vinyals} & 73.32	&68.57 & 64.37 &60.09 &58.74 &54.76 &53.15 &50.54 &49.48	&48.18	&46.86 &26.46 \\
CEC \cite{zhang2021few} & 75.85	&71.94 & 68.50 &62.50 &62.43 &58.27 &57.73 &55.81 &54.83	&53.52	&52.28 &23.57\\


NC-FSCIL \cite{yang2023neural} & 80.45 &75.98 &72.30 &70.28 &68.17 &65.16 &64.43 &63.25 &60.66 &60.01 &59.44 &21.01\\

M2SD \cite{lin2024m2sd}& 81.49 &76.67 &73.58 &68.77 &68.73 &65.78 &64.73 &64.03 &62.70 &62.09 &60.96 &20.53\\

\hline
 
\textbf{SAAN} & 78.42	&74.94	&71.54	&67.17	&66.67	&63.38	&62.43	&61.27	&60.49	&59.55	&58.55 &19.87 \\
FACT$^{\ddagger}$\cite{zhou2022forward} & 77.66	&73.70	&70.23	&65.97	&65.24	&61.92	&61.20	&59.51	&57.81	&57.35	&56.11 &21.55\\

\textbf{FACT+SAAN} & 78.21	&75.12	&72.44	&67.97	&67.19	&64.38	&63.83	&63.29	&62.17	&61.77	&60.22 &17.99\\

SAVC$^{\ddagger}$\cite{song2023learning} & 82.21	&78.30	&75.25	&70.65	&70.11	&66.96	&66.36	&65.32	&63.61	&63.86	&62.14 &20.07\\
\textbf{SAVC+SAAN} & \textbf{82.55} & \textbf{79.81} & \textbf{77.04} & \textbf{72.39} & \textbf{72.42} & \textbf{68.97} & \textbf{68.83} & \textbf{67.62} & \textbf{66.25} & \textbf{65.94} & \textbf{65.22} & \textbf{17.33} \\
\bottomrule
\end{tabular}
}
\label{tab:cub}
\end{table}

\begin{table*}
\centering
\caption{Performance comparison on \emph{mini}ImageNet dataset. *: FSCIL results implemented by \cite{tao2020few}. $\ddagger$:reproduced under the FSCIL setting with the source code. Drop: The drop in Acc. of the last session relative to session 0.}
\setlength{\tabcolsep}{1mm}
\resizebox{\textwidth}{!}{%
\begin{tabular}{ccccccccccc}
\toprule
\multirow{2}{*}{Method} & \multicolumn{9}{c}{Acc. in each session (\%)$\uparrow$} & \multirow{2}{*}{Drop$\downarrow$}\\
\cline{2-10} & 0 & 1 & 2 & 3 & 4 & 5 & 6 & 7 & 8  \\
\hline
Finetune*& 61.31 & 27.22 & 16.37 & 6.08 & 2.54 & 1.56 & 1.93 & 2.60 & 1.40 & 59.91 \\
iCaRL* \cite{rebuffi2017icarl} & 61.31 & 46.32 & 42.94 & 37.63 & 30.49 & 24.00 & 20.89 & 18.80 & 17.21 & 44.10 \\
EEIL* \cite{castro2018end} &61.31 &46.58 &44.00 &37.29 &33.14 &27.12 &24.10 &21.57 &19.58&41.73\\
TOPIC \cite{tao2020few} & 61.31 & 50.09 & 45.17 & 41.16 & 37.48 & 35.52 & 32.19 & 29.46 & 24.42 & 36.89 \\
Decoupled-Cosine$^{\ddagger}$ \cite{NIPS2016_vinyals} & 70.10 & 64.97 & 61.01 & 57.76 & 54.73 & 51.93 & 49.27 & 47.37 & 45.77 & 24.33 \\
CEC \cite{zhang2021few} & 72.00	&66.83 & 62.97 &59.43 &56.70 &53.73 &51.19 &49.24 &47.63&24.37\\


NC-FSCIL \cite{yang2023neural}& \textbf{84.02} &76.80 &72.00 &67.83 &66.35 &64.04 &61.46 &59.54 &58.31 &25.71\\

M2SD \cite{lin2024m2sd}& 82.11 &\textbf{79.92} &\textbf{75.44} &\textbf{71.31} &{68.29} &64.32 &61.13 &58.64 &56.51 &25.60\\

\hline 

\textbf{SAAN} & 74.90 & 69.97 &65.59 & 61.93 & 58.91 & 55.82 & 53.16 & 50.97 &49.98& 24.92 \\

FACT$^{\ddagger}$\cite{zhou2022forward} & 75.65 & 70.88 & 66.14 & 62.33 & 58.96 & 55.65 & 52.73 & 50.59 & 48.60 & 27.05 \\

\textbf{FACT+SAAN} & 76.11 & 71.98 & 67.21 & 64.90 & 60.12 & 58.83 & 56.28 & 54.22 & 52.11 & 24.00 \\

SAVC$^{\ddagger}$\cite{song2023learning} & 80.93 & 75.77 & 71.96 & 68.44 & 65.71 & 62.22 & 59.12 & 57.18 & 55.98 & 24.95 \\

\textbf{SAVC+SAAN} & {80.97} & {78.28} & {74.43} & {71.03} & \textbf{68.39} & \textbf{65.22} & \textbf{62.42} & \textbf{60.46} & \textbf{59.31} & \textbf{21.66} \\

\bottomrule
\end{tabular}
}
\label{tab:miniimagenet}
\end{table*}

\begin{table*}[!ht]
\centering
\caption{Performance comparison on CIFAR100 dataset. *: FSCIL results implemented by \cite{tao2020few}. $\ddagger$:reproduced under the FSCIL setting with the source code. Drop: The drop in Acc. of the last session relative to session 0.}
\setlength{\tabcolsep}{1mm}
\resizebox{\textwidth}{!}{%
\begin{tabular}{ccccccccccc}
\toprule
\multirow{2}{*}{Method} & \multicolumn{9}{c}{Acc. in each session (\%)$\uparrow$} & \multirow{2}{*}{Drop$\downarrow$}\\
\cline{2-10} & 0 & 1 & 2 & 3 & 4 & 5 & 6 & 7 & 8  \\
\hline
Finetune*& 64.10 & 39.61 &  15.37 & 9.80 & 6.67 & 3.80 &  3.70 & 3.14 & 2.65  & 61.45\\
iCaRL* \cite{rebuffi2017icarl} & 64.10 & 53.28 & 41.69 &34.13 & 27.93 &25.06 &20.41 &15.48 &13.73  &50.37 \\
EEIL* \cite{castro2018end} &64.10 &53.11 &43.71 &35.15 &28.96 &24.98 &21.01 &17.26 &15.85 &48.25\\
TOPIC \cite{tao2020few} & 64.10 & 55.88& 47.07 &45.16& 40.11& 36.38& 33.96& 31.55 &29.37 &34.73 \\
Decoupled-Cosine$^{\ddagger}$ \cite{NIPS2016_vinyals} & 74.02 & 68.71& 64.24 &60.25& 56.91& 53.87& 51.57& 49.57 &47.54 &26.48 \\
CEC \cite{zhang2021few} & 73.07	&68.88 & 65.26 &61.19 &58.09 &55.57 &53.22 &51.34 &49.14&\textbf{23.93}\\


NC-FSCIL \cite{yang2023neural}& \textbf{82.52} &\textbf{76.82} &\textbf{73.34} &\textbf{69.68} &\textbf{66.19} &\textbf{62.85} &\textbf{60.96} &\textbf{59.02} &\textbf{56.11} &26.41\\


\hline 

\textbf{SAAN} & 78.58 & 73.02 &68.99 & 65.19 & 61.90 & 59.82 & 56.99 & 54.74 &53.98& 24.60 \\

FACT$^{\ddagger}$\cite{zhou2022forward} & 79.20 & 73.05 & 69.04 & 64.89 & 61.44 & 58.82 & 56.48 & 54.20 & 51.93  & 27.27 \\

\textbf{FACT+SAAN} & 78.11 & 72.98 & 69.21 & 64.90 & 62.12 & 59.83 & 57.28 & 54.22 & 52.11 & 26.00 \\

SAVC$^{\ddagger}$\cite{song2023learning} & 78.60 & 73.42 & 68.89 & 64.63 & 61.34 & 58.32 & 56.14 & 53.85 & 51.64	 & 26.96 \\

\textbf{SAVC+SAAN} & {79.43} & {74.02} & {69.90} & {65.89} & {62.90} & 60.18 & {58.27} & {56.41} & {54.22} & 25.21\\

\bottomrule
\end{tabular}
}
\label{tab:cifar}
\end{table*}

\subsection{Performance on Conventional Benchmarks}
\label{sec:Performance on Conventional Benchmarks}
We compare SAAN and methods embedded with SAAN against classical CIL method, $i.e.$, iCaRL \cite{rebuffi2017icarl}, EEIL \cite{castro2018end}, incremental-trainable FSCIL method, $i.e.$, TOPIC \cite{tao2020few}, frozen paradigm baseline method, Decoupled-Cosine \cite{NIPS2016_vinyals}, and incremental-frozen FSCIL methods, $i.e.$, CEC \cite{zhang2021few}, FACT \cite{zhou2022forward}, SAVC \cite{song2023learning}, NC-FSCIL \cite{yang2023neural}, and M2SD \cite{lin2024m2sd}.
Among them, FACT, SAVC, and MS2D are virtual class methods, while NC-FSCIL is a fixed prototype method.
Additionally, we present a naive model, referred to as Fine-tune, which refers direct fine-tuning of models using limited data of novel classes.
We provide a brief introduction to the comparison methods for further analysis in \ref{sec:intro_methods}, and show the detailed experiment results in Tab.\ref{tab:cub}, \ref{tab:miniimagenet}, \ref{tab:cifar}. 
Here, the accuracy of the final round and the drop in accuracy from the 0-th round to the final round are used as evaluation metrics.


Compared to Decoupled-Cosine (frozen paradigm baseline), our method exhibits an average improvement over 7.44\% across the three datasets in the final session, highlighting the importance of prior spatial allocation.
It can be observed that SAAN surpasses FACT on all three datasets, indicating that our method achieves the performance of SOTA models.
FACT occupies the sample space with virtual classes, but new classes may not appear within the space occupied by these virtual classes.
SAAN further aligns the virtual classes semantically with the new classes, addressing the limitations of FACT, which allows SSAN to achieve better performance.

SAAN does not outperform SAVC, primarily because SAVC not only uses the virtual class method but also incorporates supervised contrastive learning.
Based on the self-supervised MoCo \cite{he2020momentum} framework, it significantly increases both data volume and training capacity with additional supervisory signals, yielding highly effective results.
{The generalization capability provided by contrastive learning is crucial for more complex datasets. However, on simpler datasets like CIFAR100, SAAN already outperforms SAVC. This is because SAAN requires generalized features. In simple datasets, the model can easily obtain relatively generalized features, but in more challenging datasets, additional assistance, such as supervised contrastive learning, is needed.}

As a plug-in, SAAN is compatible with both FACT and SAVC, demonstrating excellent generalization with virtual class methods.
SAVC + SAAN achieves SOTA accuracy in the final round, with improvements of 2.58\%, 3.08\%, and 3.33\% on CIFAR100, CUB200, and \emph{mini}ImageNet, respectively.
With the help of SAAN, the average forgetting rates of FACT and SAVC decreased by 2.63\% and 2.59\%, respectively, across the three datasets. This indicates that allocating space for new classes to reduce the movement of old classes can effectively lower the forgetting rate.
SAAN helps SAVC achieve additional improvements, by further refining the spatial positioning of SAVC's semantic virtual classes, and by upgrading the NCM classifier used in SAVC to better leverage the angle and norm information from limited samples.
Moreover, the methods combined with SAAN show a relatively smaller decrease in accuracy during continual learning, indicating that SAAN enhances the model's ability for incremental learning, rather than solely improving the recognition of base classes.

{SAAN shows significant performance advantages on CUB200 and achieves good results on \emph{mini}ImageNet. However, it performs worse than NC-FSCIL on CIFAR100. One possible explanation is that in SAAN, prototype allocation considers semantic information of the samples, and prototypes are adjustable. This gives SAAN good performance on fine-grained datasets like CUB200, where certain bird species have high similarity, and the similarity between different classes is more important to consider. In contrast, on datasets like CIFAR100, where there is a larger difference between categories, the differences between each category are relatively large, so the importance of carefully adjusting and allocating prototypes may be reduced. Instead, the random allocation and fixed strategy in NC-FSCIL ensure the stability of prototypes during the incremental process, leading to the best performance.}


\begin{table}[ht]
\centering

\caption{{The experimental results under open-ended and imbalanced conditions on CompCars dataset. LA: the Last session Accuracy(\%), AA: the Average Accuracy (\%), $\Delta_{last}$: Relative improvements of the last sessions compared to the Decoupled-Cosine, $\Delta_{average}$: Relative improvements of the last sessions compared to the Decoupled-Cosine, Drop: The drop in accuracy of the last session relative to session 0.}}
\renewcommand{\arraystretch}{1.3}
\resizebox{0.7\textwidth}{!}{
\label{comparison-table}
\begin{tabular}{lccccc}
\toprule
Method & LA $\uparrow$& $\Delta_{last}$ & AA $\uparrow$ & $\Delta_{average}$ & Drop $\downarrow$ \\
 \midrule
Decoupled-Cosine\cite{NIPS2016_vinyals}  & 56.24  & --   & 69.23 & -- &29.76 \\
\textbf{SAAN}    & 62.27 & +6.03   & 76.24 & +7.01 & 31.38 \\
FACT\cite{zhou2022forward}    & 61.55 & +5.31   & 75.23 & +6.00 &31.23 \\
\textbf{FACT+SAAN}& 63.59 & +7.35   & 76.52 & +7.29 &29.60\\
SAVC \cite{song2023learning}   & 70.76 & +14.52   & 81.32 & +12.09 & 24.24 \\
\textbf{SAVC+SAAN}& \textbf{73.59} & \textbf{+17.35} & \textbf{83.33} & \textbf{+14.10} &\textbf{21.93} \\

\bottomrule
\end{tabular}
}
\label{tab:open-ended}
\end{table}

\subsection{{Comparisons in Open-Ended and Imbalanced Scenarios}}
\label{sec:Comparisons in Open-Ended and Imbalanced Scenarios}


{To better simulate real-world class-incremental task scenarios, we conduct the experiments under open-ended and imbalanced conditions. Unlike the experiments in Sec. \ref{sec:Performance on Conventional Benchmarks}, we use the CompCars dataset, a fine-grained vehicle dataset with 431 car categories. Among them, 231 classes are used for the base session, while 200 classes are allocated for the incremental sessions. To simulate long-term few-shot class-incremental tasks, we extend the number of sessions to 21, including 20 incremental sessions. Additionally, the number of categories in each incremental session is sampled from a Gaussian distribution with a mean of 10 and a variance of 5, rather than being fixed at 10 classes. Similarly, the number of samples per category is sampled from a Gaussian distribution with a mean of 5 and a variance of 2, rather than being fixed at 5 samples. This setup is designed to simulate the class and sample imbalance characteristics in the real world. Due to the randomness in the task setup, we conduct 3 repeated experiments based on this configuration, and all experimental data are presented as the average of repeated experiments.}

{We perform experiments on Decoupled-Cosine, FACT, SAVC and SAAN, and integrate our method as a plug-in into FACT and SAVC, like Sec. \ref{sec:Performance on Conventional Benchmarks}. To ensure fairness, all methods are implemented using the same random seeds. We show the experiment results in Tab. \ref{tab:open-ended}, and the detailed results are provided in \ref{sec:Open-Ended Detailed Results.}. Here, the Last session Accuracy (LA), the Average Accuracy of all sessions (AA) and The drop in accuracy of the last session relative to session 0 (Drop) are used as evaluation metrics.}

{The experimental results are largely consistent with those in Sec. 
\ref{sec:Performance on Conventional Benchmarks}, particularly resembling the outcomes on CUB200 in Tab. \ref{tab:cub}. This aligns with our expectations, as both being fine-grained datasets, CompCars poses similar challenges to these methods as CUB200 does. SAAN shows an improvement of 6.03\% in LA and 7.01\% in AA over decoupled-cosine (the frozen paradigm baseline), and an enhancement of 0.72\% in LA and 1.01\% in AA over FACT, indicating that our method can achieve the performance of SOTA models when used independently, and remains effective in open-ended and imbalanced scenarios. SAAN does not outperform SAVC, which is also in line with our previous analysis. Compared to coarse-grained datasets with distinct inter-class differences such as CIFAR100, models require greater generalization capability on fine-grained datasets, and the supervised contrastive learning used in SAVC can well compensate for this. Therefore, when we integrate SAAN as a plugin into SAVC, the model's performance significantly surpasses that of either SAAN or SAVC alone, with a 2.83\% improvement in LA and a 2.01\% improvement in AA over SAVC, achieving the highest performance and the lowest forgetting rate. SAAN enhances the performance of SAVC through two key refinements: it meticulously adjusts the spatial arrangement of SAVC's semantic virtual categories and enhances the NCM classifier within SAVC to more effectively utilize the angular and magnitude data derived from a constrained set of samples. Embedding SAAN into FACT also boosts FACT's performance. This demonstrates that SAAN remains effective as a plugin in open-ended and imbalanced scenarios.}

\begin{table}
\centering
\caption{Ablation studies on CIFAR100 dataset. $\Delta_{last}$: Relative improvements of the last sessions compared to the baseline.}
\resizebox{\textwidth}{!}{
\begin{tabular}{cccccccccccccc}
\toprule
\multirow{2}{*}{$\mathcal{L}_1$} & \multirow{2}{*}{$\mathcal{L}_2$} &\multirow{2}{*}{\textbf{2S}}& \multirow{2}{*}{\textbf{ND}}&\multicolumn{9}{c}{Acc. in each session (\%)$\uparrow$} & \multirow{2}{*}{$\Delta_{last}$}\\
\cline{5-13} & & & & 0 & 1 & 2 & 3 & 4 & 5 & 6 & 7 & 8 \\
\hline 
 && & & 74.02	&68.71	&64.24	&60.25	&56.91	&53.87	&51.57	&49.57	&47.54 & --  \\
 $\checkmark$& & & &  77.05 & 71.66 & 67.27 & 63.56 & 60.30 & 57.55 & 55.12 & 53.08 & 50.95 & +3.41 \\
 $\checkmark$& $\checkmark$& && 77.63 & 72.23 & 68.22 & 64.28 & 60.94 & 58.12 & 55.70 & 53.58 & 51.55  & +4.01 \\
 $\checkmark$& $\checkmark$ & $\checkmark$ && 78.08	&72.45	&68.42	&64.45	&61.18	&58.39	&55.91	&53.79	&51.73 & +4.19 \\ 
 $\checkmark$& $\checkmark$ && $\checkmark$ & 78.10	&72.79	&68.73	&65.05	&61.63	&59.63	&56.74	&54.58	&52.72 & +5.18 \\
 $\checkmark$& $\checkmark$ &$\checkmark$& $\checkmark$ & 78.58	&73.02	&68.99	&65.19	&61.90	&59.82	&56.99	&54.74	&52.98 & +5.44 \\
\bottomrule
\end{tabular}
}
\label{tab:ablation}
\end{table}

\begin{table}
\centering
\caption{{ANJ ablation study on CUB200 dataset. }}
\resizebox{\textwidth}{!}{
\begin{tabular}{ccccccccccccc}
\toprule
\multirow{2}{*}{Distribution} &  \multirow{2}{*}{Shared Distribution}&\multicolumn{11}{c}{Acc. in each session (\%)$\uparrow$} \\
\cline{3-13}  & & 0 & 1 & 2 & 3 & 4 & 5 & 6 & 7 & 8 & 9 & 10 \\
\hline 

\multirow{2}{*}{Normal} & $\times$ &78.19	&73.84	&70.29	&66.36	&65.68	&62.11	&61.55	&60.10	&59.69	&58.20	&57.56   \\
 & $\checkmark$ &78.19	&74.56	&71.14	&66.79	&66.29	&63.02	&62.03	&60.87	&60.01	&59.15	&58.18   \\
 \hline
 \multirow{2}{*}{Log-Normal} & $\times$ & 78.42	&74.03	&70.63	&66.39	&65.90	&62.52	&61.82	&60.44	&59.66	&58.72	&57.72  \\
 & $\checkmark$ & 78.42	&74.94	&71.54	&67.17	&66.67	&63.38	&62.43	&61.27	&60.49	&59.55	&58.55   \\ 
 \hline
 \multirow{2}{*}{Weibull} & $\times$ & 78.43	&74.48	&70.90	&66.75	&65.94	&62.80	&61.63	&60.18	&59.68	&58.96	&57.70   \\
 & $\checkmark$  &78.43	&74.98	&71.58	&67.30	&66.62	&63.48	&62.47	&61.16	&60.51	&59.71	&58.65   \\
 \hline
 \multirow{2}{*}{Patero} & $\times$ & 76.44	&72.03	&68.33	&64.30	&63.85	&60.09	&60.02	&58.62	&57.61	&56.86	&55.46    \\
  & $\checkmark$ & 76.44	&72.30	&68.62	&64.70	&64.14	&60.45	&60.48	&58.95	&57.99	&57.19	&55.76   \\
\bottomrule
\end{tabular}
}
\label{tab:ablation_anj}
\end{table}

\subsection{Ablation Studies}
\label{sec:Ablation}
We conduct ablation experiments to verify the importance of the components of our method.
Building upon the incremental freezing framework, we set Decoupled-Cosine as the baseline and then gradually add $\mathcal{L}_1$, $\mathcal{L}_2$, 2SNCM (\textbf{2S}), and Norm Distribution (\textbf{ND}) to observe their respective contributions.
The results on the CIFAR100 dataset are presented in Tab. \ref{tab:ablation}.
{}

The experiments reveal that $\mathcal{L}_1$ has the most significant impact, with a 3.41\% improvement when combined with the baseline, highlighting the importance of embedding space allocation.
$\mathcal{L}_2$ aims to increase inter-class distances, resulting in a 4.01\% improvement.
2SNCM achieves a modest improvement of 4.19\%, indicating a limited enhancement.
This indicates that 2SNCM can achieve more reasonable representative points simply by normalizing and averaging the samples in the base round, ultimately leading to a certain performance improvement.
To further verify the effectiveness of 2SNCM, we present multiple repeated experiments on three datasets in \ref{sec:2SNCM}.
Norm Distribution achieves an improvement of 5.18\%, demonstrating the effectiveness of introducing norm information to the classifier.
The ANJ, composed of 2SNCM and Norm Distribution, ultimately achieves a performance improvement of 5.44\%.
The components in ANJ do not participate in model training but construct a more accurate classifier based on the feature extractor, with minimal computational cost.
In summary, the spatial allocation strategy of CCSA is highly beneficial for FSCIL, while ANJ aids classification by making more comprehensive and effective use of the features.

{To verify the rationality of the distribution assumption in ANJ and the effectiveness of the shared distribution technique, we conduct an ablation study on ANJ alone, as shown in Tab. \ref{tab:ablation_anj}. The log-normal, normal, and Weibull distributions can effectively represent feature norms, leading to a certain improvement in model performance. However, the Pareto distribution, a heavy-tailed distribution, rendered ANJ ineffective due to its significant difference from the norm distribution. Among all distributions, the Weibull and log-normal distributions performed the best. The log-normal distribution has the advantage of simple parameter estimation, whereas the Weibull distribution requires storing data for distribution fitting. Without adopting the shared distribution technique, the incremental data becomes too sparse, leading to a significant discrepancy between the estimated distribution and the true distribution. This issue is particularly severe for the Weibull distribution, which requires a sufficient amount of data for proper fitting. Regardless of the chosen distribution, without the shared distribution technique, the performance deteriorates significantly. By leveraging the shared distribution technique, the model can better identify the session to which a sample belongs, thereby indirectly improving the final classification performance.}

\begin{figure}
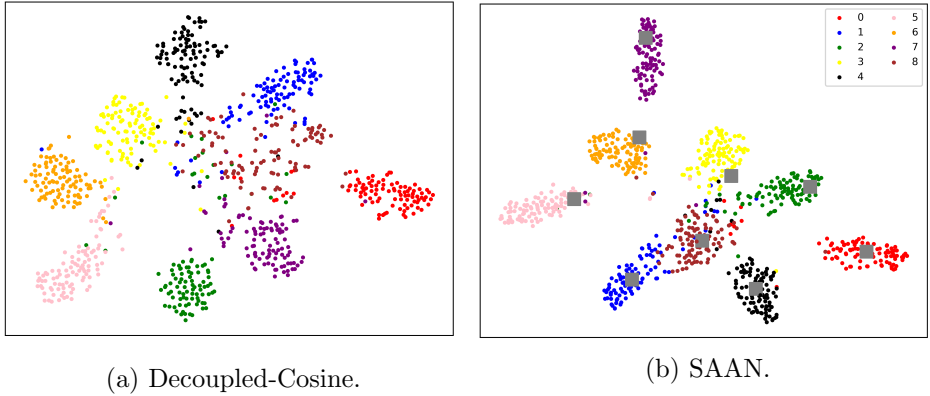

  \centering
  \begin{subfigure}{0.45\textwidth}
    \includegraphics[width=\linewidth, trim=600 300 400 400, clip]{figures/t\_sne\_dc.png}
    \caption{Decoupled-Cosine.}
    
  \end{subfigure}
  \begin{subfigure}{0.45\textwidth}
    \includegraphics[width=\linewidth, trim=450 300 300 300, clip]{figures/t\_sne\_ccsa.png}
    \caption{SAAN.}
    \label{Fig:T-SNE_CCSA}
  \end{subfigure}
  \caption{The t-SNE visualization on CIFAR100 dataset of the embeddings. Classes 0-5 represent the base classes while classes 6-8 represent the novel classes. The squares represent the centers set by SAAN.}
  \label{Fig:T-SNE}
\end{figure}

\subsection{Numerical Analysis}
\label{sec:normal_test}
\label{sec:parameter_sensitivity}
\textbf{Visualization of Embedding Separation.}
We randomly choose 6 classes from base classes and 3 classes from novel classes respectively and visualize the embeddings of them on the CIFAR100 dataset with t-SNE \cite{van2008visualizing} in Fig. \ref{Fig:T-SNE}.
Firstly, it can be observed that the relative positions of class embeddings shift compared to Decoupled-Cosine.
In Decoupled-Cosine, the purple, brown, and red classes are adjacent, while in SAAN, the green, black, and red classes are adjacent.
This is because the allocation of class centers affects the positions of class means.
During the class increment process, in SAAN, the purple and brown new classes are assigned to two different reserved spaces, whereas in Decoupled-Cosine, the purple and brown classes are assigned to adjacent spaces and are very close to each other.
This makes it difficult to distinguish between the purple and brown classes in Decoupled-Cosine. While in SAAN, it is easier to identify them.
Additionally, in SAAN, the inter-class distance is larger, while the intra-class distance is smaller.


In Fig. \ref{Fig:T-SNE_CCSA}, it can be observed that the sample embeddings are distributed around the centers we set, indicating that the limiting effect of the centers is effective.
There is some deviation between these centers and the class mean, which is due to the influence of cross-entropy.

\begin{figure}
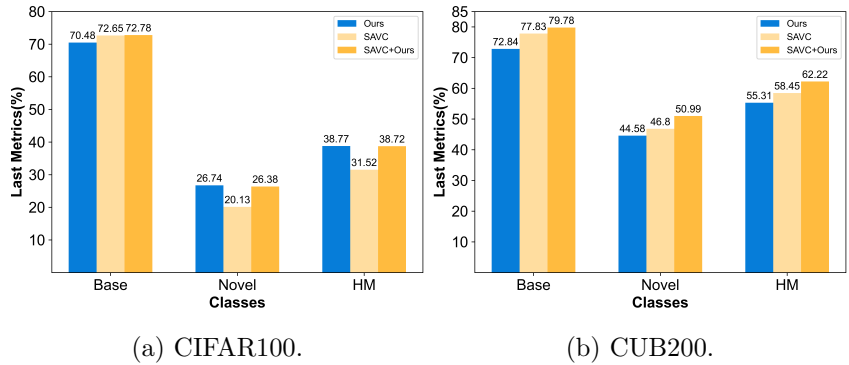

  \centering
  \begin{subfigure}{0.4\textwidth}
    {\includegraphics[width=\linewidth,trim=10 1 10 10,clip]{figures/HM\_CIFAR.png}}
    \caption{CIFAR100.}
  \end{subfigure}
  \begin{subfigure}{0.4\textwidth}
   {\includegraphics[width=\linewidth,trim=10 1 10 10,clip]{figures/HM\_CUB200.png}}
    \caption{CUB200.}
  \end{subfigure}
  
  \caption{Accuracy of base and novel classes at the last session.}
  \label{fig:hm}
\end{figure}
\textbf{Performance on Novel and Base Classes.} 
Solely based on the accuracy over all classes, we cannot evaluate how much new knowledge the model has learned and how much old knowledge it has forgotten.
We report the accuracy of the base classes and the novel classes as well as harmonic mean in Fig. \ref{fig:hm}.
With the help of SAAN, SAVC's accuracy for novel classes improves by 7.20\% and 3.76\% in CIFAR100 and CUB200 respectively.
The accuracy improvement for old classes is very limited compared to that for new classes.
SAAN obviously enhances the compatibility of the model to new classes verifying that limiting the space occupation of old classes and allocating space for new classes are conducive to the learning of new classes.
\begin{figure}
    \centering
    \subcaptionbox{\label{fig:alpha_beta}}
    {\includegraphics[width=0.45\textwidth,trim=0 0 0 0,clip]{figures/alpha\_beta}}
    \subcaptionbox{\label{fig:eta_lambda}}
    {\includegraphics[width=0.45\textwidth,trim=0 0 0 0,clip]{figures/eta\_lambda}}

    \caption{{The parameter grid search for the last session accuracy of SAVC + SAAN on the CUB200 dataset. (a) The weights of $\mathcal{L}_1$ and $\mathcal{L}_2$. (b) The center moving rate and its decay rate.}}
    \label{fig:grid_parameter_search}
\end{figure}
\begin{figure}
    \centering
    \subcaptionbox{\label{fig:parameter_a}}
    {\includegraphics[width=0.3\textwidth,trim=0 0 0 0,clip]{figures/search\_alpha.png}}
    \subcaptionbox{\label{fig:parameter_b}}
    {\includegraphics[width=0.3\textwidth,trim=0 0 0 0,clip]{figures/search\_eta.png}}
    \subcaptionbox{\label{fig:parameter_C}}
    {\includegraphics[width=0.3\textwidth,trim=0 0 0 0,clip]{figures/search\_C.png}}
    {\includegraphics[width=0.6\textwidth,trim=0 25 0 20,clip]{figures/parameter\_legend.png}}
    
    \caption{{Last session accuracy of SAVC + SAAN with different hyper-parameters on three datasets. (a) $\alpha$ varies while $\eta$ is fixed and the ratio of $\alpha$ to $\beta$ is 5. (b) $\eta$ varies while $\alpha$ is fixed, and $\lambda$ is 0.1. (c) $C$ varies.}}
    \label{fig:parameter_search}
\end{figure}

\textbf{Parameter Sensitivity.}
{We conduct a search and discussion on the loss weights $\alpha$, $\beta$, compression coefficient $C$, as well as the center moving rate $\eta$ and its decay rate $\lambda$. The grid search results for cosine center loss weights and center moving rate with its decay rate on CUB200 are shown in Fig. \ref{fig:grid_parameter_search}. The parameter analysis across three datasets is presented in Fig. \ref{fig:parameter_search}. As $\alpha$ and $\beta$ increase, the effect of CCSA becomes stronger, leading to improved model performance, which reaches its optimum when $\alpha$ is 2 and $\beta$ is 0.4. $\mathcal{L}_1$ serves to push away from other centers, while $\mathcal{L}_2$ pulls closer to the corresponding center. If $\beta$ is too large and $\alpha$ is too small, the pushing effect will be dominant, while the more crucial pulling effect will be weakened, leading to a negative impact. For all three datasets, the optimal value of \(C\) is consistently 0.005. When \(C\) exceeds 0.05, the influence of norm logits becomes overly dominant, rendering the more important angular logits ineffective, which leads to a sharp performance drop.}

{It can be observed that the center moving rate and decay rate are relatively insensitive, as values ranging from 0.25 to 2 all have a positive effect. However, considering two special cases: when the center moving rate is 0, it represents fixed centers; when the center moving decay rate is 1, it means that the centers are continuously updated without playing a role in spatial allocation, only contributing to intra-class cohesion. Fixed centers perform better than the baseline but are inferior to the learnable SAAN. This indicates that spatial allocation is beneficial for FSCIL, and learnable centers achieve better allocation performance. Continuously updating centers perform almost the same as the baseline, failing to achieve spatial allocation. This highlights the importance of spatial allocation, as merely increasing inter-class distance has limited effectiveness.}




\begin{figure}
  \centering
  \begin{subfigure}{0.35\textwidth}
    \includegraphics[width=\linewidth, trim=20 5 45 30, clip]{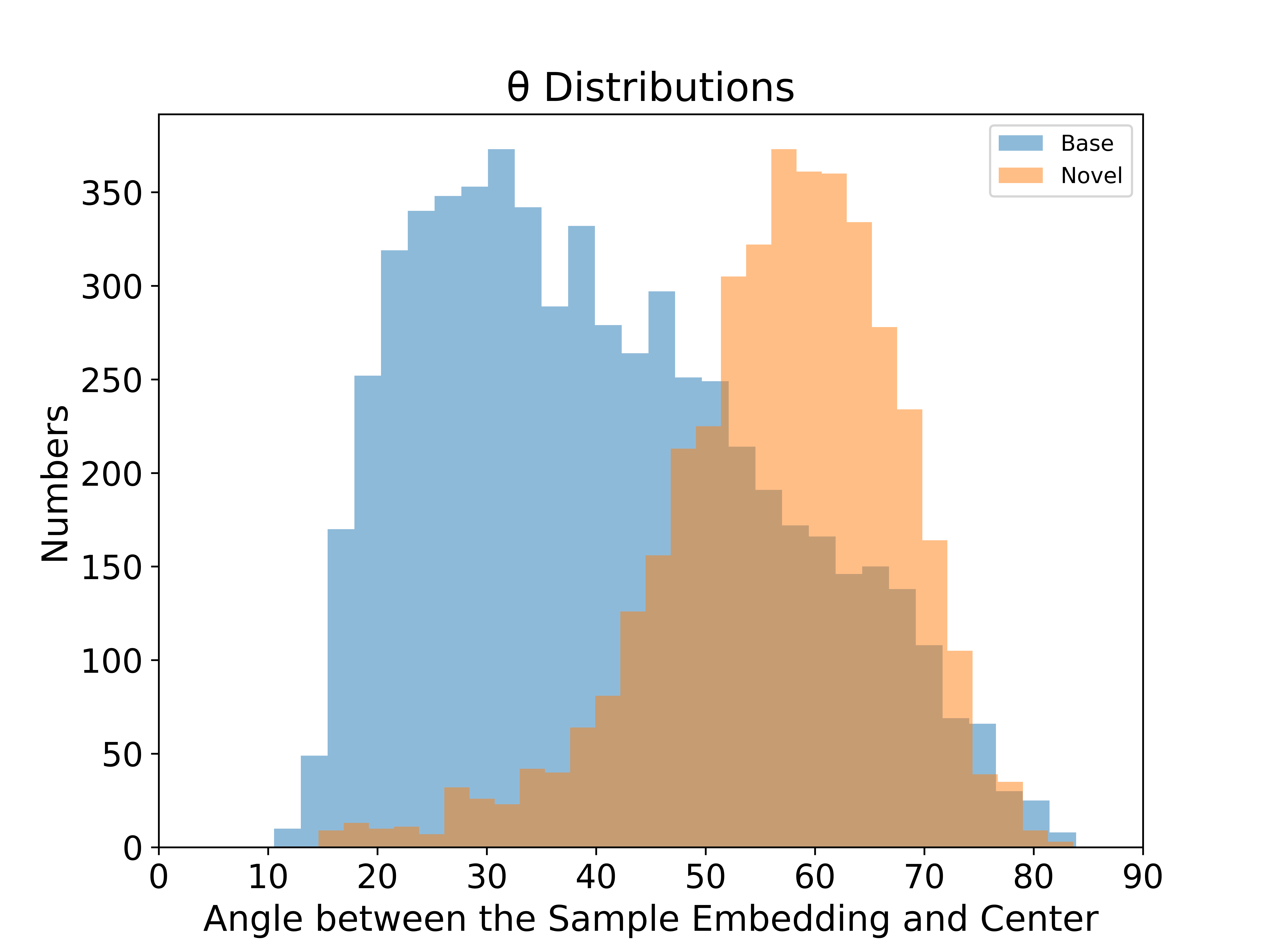}
    \caption{SAVC}
  \end{subfigure}
  \begin{subfigure}{0.35\textwidth}
    \includegraphics[width=\linewidth, trim=20 5 45 30, clip]{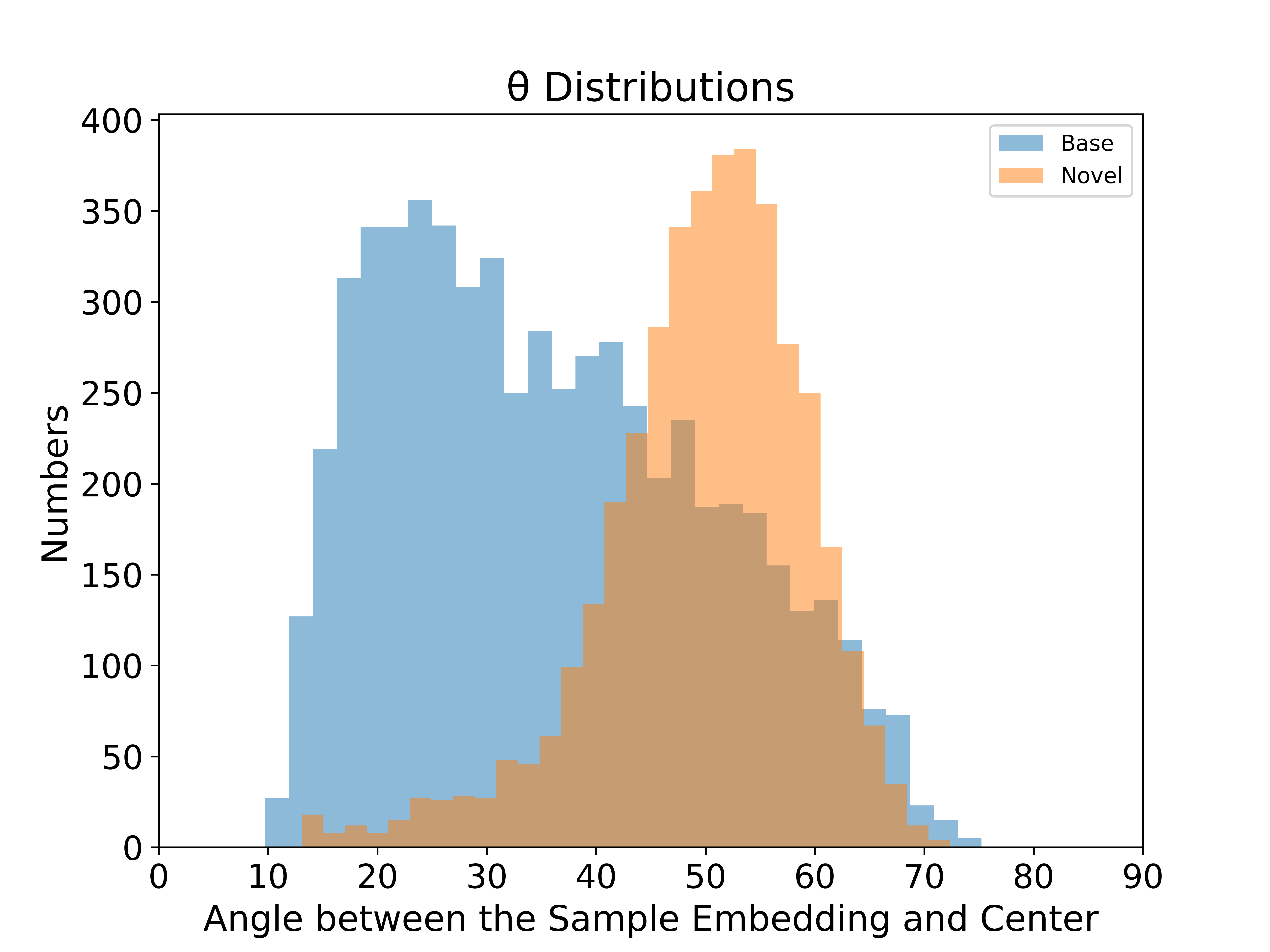}
    \caption{SAVC+SAAN}
  \end{subfigure}
  \caption{The angle distributions between the sample embeddings and their corresponding class centers on \emph{mini}ImageNet dataset. Blue represents base classes, and orange represents novel classes.}
  \label{Fig:angle}
\end{figure}

\textbf{Angle Distribution from the Centers.}
We present the angle of each sample relative to its center of SAVC and SAVC+SAAN, as shown in Fig. \ref{Fig:angle}.
It can be observed that the mean angle for novel classes of SAVC is approximately 60 degrees, while the mean angle of SAVC+SAAN is around 50 degrees, with a noticeably smaller maximum angle compared to SAVC.
The results on base classes are similar.
This implies that SAVC+SAAN exhibits a more compact intra-class structure, by using our cosine center loss.

\begin{figure}
  \centering
  \includegraphics[width=0.45\linewidth, trim=10 5 40 40, clip]{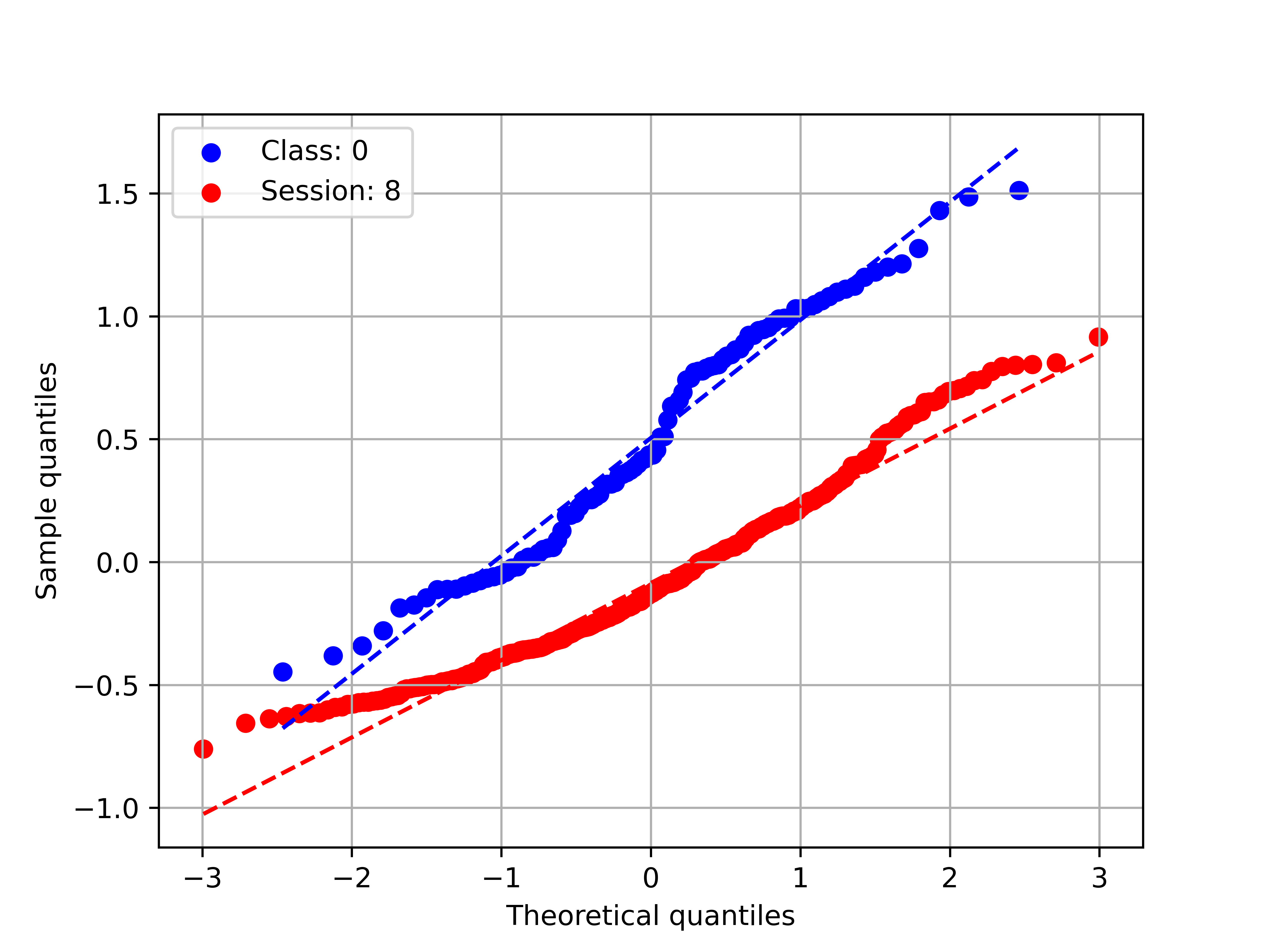}
  \caption{Quantile-Quantile plot of embedding log norms on \emph{mini}ImageNet dataset.}
  \label{fig:qq}
\end{figure}

\textbf{Normal Test.}

We present the Quantile-Quantile plot of the log norms of sample embeddings in Fig. \ref{fig:qq}, where the horizontal axis represents theoretical quantities and the vertical axis represents sample quantities.
Due to the granularity of the base session distribution being categories, we randomly select one category.
During the incremental sessions, we select one session due to distribution sharing.
It can be observed that the data points are nearly aligned along a straight line, indicating that the sample log norms can be considered to follow a normal distribution.
In session 8, the data points slightly deviate above the straight line, indicating that the shared norm distribution exhibits a long-tail behavior.


\section{Conclusion}


We summarize two challenges present in the FSCIL problem: (i) Due to the CIL problem setup, current classes occupy the entire feature space, which is detrimental to learning new classes. (ii) Due to the FSL problem setup, the small number of samples in incremental rounds is insufficient for fully training. To address these two issues and improve upon previous methods, we propose the SAAN framework. In response to challenge (i), SSAN sets the orthogonal class centers to partition the embedding space, and guides sample embeddings towards their corresponding class center, thus providing reserved space for new classes. Experiments show that under the effect of spatial allocation, the model performance improves significantly, highlighting the importance of spatial allocation. Additionally, the spatial allocation and virtual class reservation methods are compatible and jointly reserve space for new classes. In response to challenge (ii), SAAN establishes a more precise decision boundary by integrating 2SNCM and ND, making full use of angle and norm information. Experiments show that reasonably selecting representative points for classes and fully utilizing the complete embedding information to optimize the classifier, without changing the feature extraction module, still contributes to performance improvement.



\section{Acknowledgments}
This work was supported in part by the STI 2030-Major Projects of China under Grant 2021ZD0201300, and the Fundamental Research Funds for the Central Universities under Grant 2024300394.

\appendix

\section{Gradient Analysis of Cosine Center Loss}
\label{sec:Gradient_Analysis}
In this section, we provide the gradient of the proposed cosine center loss and analyze the update mechanism of the learnable centers.
Since analyzing the gradients of $\mathcal{L}_1$ and $\mathcal{L}_2$ with respect to embeddings requires deriving the gradient of the cosine similarity with respect to the vector, we give it in advance: \begin{equation}
    \frac{\partial \cos \langle \mathbf{a}, \mathbf{b}\rangle}{\partial \mathbf{a}}=\frac{\mathbf{b}}{\|\mathbf{a}\|\|\mathbf{b}\|}-\frac{\mathbf{a}}{\|\mathbf{a}\|^2} \cos \langle \mathbf{a}, \mathbf{b}\rangle,
    \label{eq:gradient_cos}
\end{equation}
where $\mathbf{a}$ and $\mathbf{b}$ are any venctors. In this paper, the norm specifically refers to the L2 norm.

\subsection{Analysis of $\mathcal{L}_1$}
First we give the negative gradient of $\mathcal{L}_1$ with respect to $\hat{\mathbf{e}}_i=\frac{\mathbf{e}_i}{\|\mathbf{e}_i \|}$ as:\begin{equation}
\begin{aligned}
    -\frac{\partial \mathcal{L}_1}{\partial \hat{\mathbf{e}}_i}&=\frac{1}{m}\hat{\mathbf{c}}_{y_i}.
\end{aligned}
\label{eq:gradient_l1}
\end{equation}
Since $\mathbf{c}_{y_i}$ is normalized upon initialization and after each updating, $\hat{\mathbf{c}}_{y_i}$ is equal to $\mathbf{c}_{y_i}$.
Eq. \ref{eq:gradient_l1} shows that the normalized sample embedding will move closer to the center according to vector addition.
Combining Eq. \ref{eq:gradient_cos} or using chain derivation role, we can further obtain the negative gradient of $\mathcal{L}_1$ with respect to ${\mathbf{e}_i}$ as:
\begin{equation}
\begin{aligned}
    -\frac{\partial \mathcal{L}_1}{\partial \mathbf{e}_i}&=-\frac{\partial \mathcal{L}_1}{\partial \hat{\mathbf{e}}_i}\frac{\partial\hat{\mathbf{e}}_i}{\partial \mathbf{e}_i}\\&=\frac{1}{m}\left( \frac{\mathbf{c}_{y_i}}{\| \mathbf{e}_i\|}-\frac{\mathbf{e}_i\cos\langle \mathbf{e}_i,\mathbf{c}_{y_i} \rangle}{{\| \mathbf{e}_i\|}^2}\right).
    \label{eq:gradient_l1_fx}
\end{aligned}
\end{equation}
Eq. \ref{eq:gradient_l1_fx} indicates that the sample embedding will move perpendicular to itself, pointing towards the center.
\(\mathcal{L}_1\) guides the samples, playing a role in spatial allocation and reducing the intra-class distance.
\(\mathcal{L}_2\) pushes the samples away from other centers, helping to increase the inter-class distance.

\subsection{Analysis of $\mathcal{L}_2$}
Likewise, we give the gradient of $\mathcal{L}_2$ with respect to $\hat{\mathbf{e}}_i$ as:\begin{equation}
    \frac{\partial \mathcal{L}_2}{\partial \hat{\mathbf{e}}_i}=\frac{1}{m}\frac{1}{|\mathcal{Y}|}\sum_{j=1}^{|\mathcal{Y}|}\mathbf{c}_{j}\mathbb{I}\left( y_i\neq j\right).
    \label{eq:gradient_l2}
\end{equation}
Eq. \ref{eq:gradient_l2} shows that the normalized embedding will move away from the mean direction of other class centers through vector subtraction.
Similarly, combining Eq. \ref{eq:gradient_cos} or using chain derivation role, we further obtain the gradient of $\mathcal{L}_2$ with respect to ${\mathbf{e}_i}$ as:
\begin{equation}
\begin{aligned}
    \frac{\partial \mathcal{L}_2}{\partial \mathbf{e}_i}&=\frac{\partial \mathcal{L}_2}{\partial \hat{\mathbf{e}}_i}\frac{\partial\hat{\mathbf{e}}_i}{\partial \mathbf{e}_i}\\&=\frac{1}{m}\frac{1}{|\mathcal{Y}|}\sum_{j=1}^{|\mathcal{Y}|}\left( \frac{\mathbf{c}_{j}}{\| \mathbf{e}_i\|}-\frac{\mathbf{e}_i\cos\langle \mathbf{e}_i,\mathbf{c}_{j} \rangle}{{\| \mathbf{e}_i\|}^2}\right)\mathbb{I}\left( y_i\neq j\right).
    \label{eq:gradient_l2_fx}
\end{aligned}
\end{equation}
Eq. \ref{eq:gradient_l2_fx} means that the sample embedding will move perpendicular to itself and point in the opposite direction to the mean of other centers.
The gradient direction is the fastest direction in which the cosine similarity between the embedding and centers of other classes decreases.

\subsection{The Center Update Mechanism.}
Another way to update the centers is to set them as learnable parameters and update them through gradient descent based on \(\mathcal{L}_1\).
Since we are concerned with the normalized center, we compute the gradient of $\mathcal{L}_1$ with respect to the normalized center as:
\begin{equation}
    \frac{\partial \mathcal{L}_{1}}{\partial \hat{\mathbf{c}}_j} = - \frac{1}{m}\sum_{i=1}^{m}\left( \hat{\mathbf{e}}_i\mathbb{I}(y_i=j)\right).
\end{equation}
Update the center through gradient descent as:
\begin{equation}
    \mathbf{c}_j=\hat{\mathbf{c}}_j - \eta \frac{\partial \mathcal{L}_{1}}{\partial \hat{\mathbf{c}}_j}.
    \label{eq:gd}
\end{equation}
It can be observed that gradient descent updates are equivalent to updates via vector addition.
That is, the centers can be adjusted to positions more aligned with the samples either through gradient descent or vector addition, reducing the unreasonable allocation from pre-assigned centers.

\section{Comparison Methods}
\label{sec:intro_methods}
To analyze the performance comparison with other methods, we provide an introduction to the comparison methods here.

• \textbf{Finetune}. It directly fine-tunes the model with limited data in incremental sessions, without addressing either the forgetting problem or overfitting.

• \textbf{iCaRL} \cite{rebuffi2017icarl}. It maintains an "episodic memory" of the exemplars, which enables one to incrementally learn new classes without forgetting old classes.

• \textbf{EEIL} \cite{castro2018end}. It evolves ensemble models gradually, adapting to new data while preserving past knowledge.

• \textbf{TOPIC} \cite{tao2020few}. It mitigates forgetting by utilizing a Neural Gas (NG) network \cite{martinetz1991neural} to preserve the topology of the feature manifold formed by different classes.

• \textbf{Decoupled-Cosine} \cite{NIPS2016_vinyals}. It is a one-shot learning method that utilizes cosine similarity to efficiently learn new classes with minimal data by decoupling the classifier from the feature extractor.

• \textbf{CEC} \cite{zhang2021few}. It introduces an auxiliary graph model that promotes the exchange of contextual information among classifiers, thereby improving the adaptation process.

• \textbf{FACT} \cite{zhou2022forward}. It pre-assigns multiple virtual prototypes and generates virtual instances via instance mixture in the embedding space, to reserve spaces for incoming new classes.

• \textbf{NC-FSCIL} \cite{yang2023neural}. It introduces a neural collapse-inspired feature-classifier alignment method for FSCIL, addressing the misalignment issue between features and classifiers in incremental learning with superior performance demonstrated on multiple datasets.


• \textbf{M2SD} \cite{lin2024m2sd}. It proposes a dual-branch structure that facilitates the expansion of the feature space and employs self-distillation to pass additional enhanced information back to the base network.

• \textbf{SAVC} \cite{song2023learning}. It introduces more semantic knowledge by imagining virtual classes which not only hold enough feature space for future updates but enable a multi-semantic aggregated inference effect.

\section{{Detailed Results of the Open-Ended Experiments}}
\label{sec:Open-Ended Detailed Results.}
{To better simulate the real-world environment of class-incremental tasks, we conduct this comparative experiment under open-ended and imbalanced conditions. The experimental setup and the analysis are elaborated in Sec. \ref{sec:Comparisons in Open-Ended and Imbalanced Scenarios}. The experimental results we obtain are presented in Tab. \ref{tab:detailed-CompCars}.}

\begin{table*}[!ht]
\centering
\caption{{Detailed results of the open-ended and imbalanced experiments on CompCars dataset.}}
\setlength{\tabcolsep}{1mm}
\resizebox{\textwidth}{!}{%
\renewcommand{\arraystretch}{1.2}
\begin{tabular}{ccccccccccccccccccccccc}
\toprule
\multirow{2}{*}{Method} & \multirow{2}{*}{Random} & \multicolumn{21}{c}{Acc. in each session (\%)$\uparrow$} \\
\cline{3-23}
& & 0 & 1 & 2 & 3 & 4 & 5 & 6 & 7 & 8 & 9 & 10 & 11 & 12 & 13 & 14 & 15 & 16 & 17 & 18 & 19 & 20 \\
\hline
\multirow{3}{*}{Decoupled-Cosine\cite{NIPS2016_vinyals}}
& rand 1 & 85.88 & 83.90 & 82.88 & 81.79 & 80.57 & 79.13 & 76.44 & 73.40 & 71.61 & 69.15 & 67.27 & 64.40 & 63.34 & 61.57 & 60.58 & 59.77 & 58.71 & 57.87 & 57.44 & 56.30 & 56.22 \\
& rand 2 & 86.26 & 85.02 & 83.89 & 81.68 & 80.69 & 79.15 & 76.53 & 74.11 & 71.93 & 69.49 & 67.27 & 64.80 & 63.56 & 61.95 & 60.94 & 59.02 & 58.58 & 57.72 & 56.96 & 56.37 & 55.56 \\
& rand 3 & 85.88 & 84.36 & 83.16 & 82.21 & 80.85 & 79.46 & 77.00 & 74.30 & 72.37 & 70.07 & 67.90 & 65.29 & 64.10 & 62.32 & 61.30 & 60.50 & 59.52 & 58.53 & 58.10 & 57.38 & 56.96 \\
\hline
\multirow{3}{*}{\textbf{SAAN}}
 & rand 1 & 93.79 & 91.89 & 90.70 & 89.80 & 88.51 & 86.98 & 84.44 & 81.53 & 79.79 & 77.41 & 75.16 & 72.23 & 71.24 & 69.12 & 68.15 & 67.06 & 66.05 & 64.96 & 64.33 & 63.37 & 62.91\\
& rand 2 & 93.40 & 91.80 & 90.45 & 88.83 & 87.77 & 85.96 & 83.08 & 80.40 & 78.36 & 75.57 & 73.44 & 70.78 & 69.55 & 67.75 & 66.56 & 64.62 & 64.09 & 63.15 & 62.29 & 61.56 & 60.82\\
& rand 3 & 93.79 & 91.95 & 90.97 & 90.07 & 88.78 & 87.31 & 84.83 & 81.97 & 80.26 & 77.58 & 75.49 & 72.58 & 71.35 & 69.28 & 68.09 & 67.05 & 66.06 & 65.01 & 64.39 & 63.50 & 63.09\\
\hline

\multirow{3}{*}{FACT\cite{zhou2022forward}}  
 & rand 1 & 92.76 & 91.00 & 89.61 & 88.74 & 87.18 & 85.73 & 83.10 & 80.07 & 78.25 & 75.64 & 73.23 & 71.03 & 69.55 & 67.72 & 66.71 & 65.74 & 64.75 & 63.80 & 63.30 & 62.48 & 62.10\\  
& rand 2 & 92.82 & 91.40 & 89.59 & 87.96 & 86.97 & 85.24 & 82.64 & 79.98 & 77.93 & 75.18 & 72.91 & 70.31 & 68.96 & 67.08 & 66.00 & 63.97 & 63.54 & 62.67 & 61.83 & 61.16 & 60.15\\  
& rand 3 & 92.76 & 90.85 & 89.81 & 88.87 & 87.43 & 85.93 & 83.31 & 80.44 & 78.79 & 76.17 & 73.76 & 71.56 & 70.06 & 68.07 & 67.00 & 66.06 & 65.13 & 64.05 & 63.54 & 62.83 & 62.40\\  
\hline

\multirow{3}{*}{\textbf{FACT+SAAN}}
& rand 1 & 93.44 & 91.44 & 89.77 & 88.92 & 87.87 & 86.27 & 84.05 & 81.41 & 79.64 & 77.16 & 75.21 & 72.90 & 71.48 & 69.49 & 68.49 & 67.75 & 66.92 & 66.00 & 65.54 & 64.78 & 64.40\\
& rand 2 & 92.54 & 90.72 & 89.53 & 88.05 & 87.07 & 85.26 & 82.76 & 80.20 & 78.23 & 75.48 & 73.70 & 71.06 & 69.83 & 68.02 & 66.92 & 64.96 & 64.55 & 63.62 & 62.76 & 62.26 & 61.46\\
& rand 3 & 93.61 & 92.19 & 90.65 & 89.85 & 88.70 & 87.05 & 84.89 & 82.48 & 80.70 & 78.43 & 76.02 & 73.74 & 72.32 & 70.32 & 69.37 & 68.32 & 67.50 & 66.49 & 65.97 & 65.56 & 64.92\\
\hline

\multirow{3}{*}{SAVC\cite{song2023learning}}
& rand 1 & 95.31 & 93.18 & 91.92 & 91.10 & 90.53 & 89.01 & 87.43 & 85.07 & 84.05 & 81.91 & 80.29 & 78.37 & 77.59 & 76.22 & 75.47 & 74.98 & 74.20 & 73.35 & 73.11 & 72.09 & 71.85\\
& rand 2 & 94.38 & 92.37 & 91.45 & 90.39 & 89.11 & 87.78 & 86.11 & 84.20 & 82.53 & 80.34 & 78.65 & 76.71 & 75.38 & 73.93 & 73.11 & 71.36 & 71.05 & 70.32 & 69.44 & 68.82 & 68.10\\
& rand 3 & 95.31 & 93.69 & 92.42 & 91.75 & 90.97 & 89.77 & 88.16 & 86.08 & 85.06 & 82.98 & 81.54 & 79.36 & 78.33 & 76.81 & 76.11 & 75.36 & 74.51 & 73.84 & 73.36 & 72.93 & 72.33\\
\hline

\multirow{3}{*}{\textbf{SAVC+SAAN}}
& rand 1 & 95.60 & 94.13 & 92.86 & 92.52 & 91.74 & 90.40 & 88.73 & 86.63 & 85.77 & 83.91 & 83.02 & 80.95 & 79.94 & 78.59 & 77.74 & 77.36 & 76.63 & 75.71 & 75.45 & 74.56 & 74.37\\
& rand 2 & 95.35 & 94.08 & 93.52 & 92.25 & 91.07 & 89.83 & 87.96 & 86.36 & 84.87 & 82.71 & 81.30 & 79.81 & 78.89 & 77.37 & 76.63 & 75.10 & 74.58 & 73.78 & 72.88 & 72.37 & 71.86\\
& rand 3 & 95.60 & 94.63 & 93.46 & 92.79 & 91.91 & 90.68 & 89.18 & 87.32 & 86.23 & 84.37 & 83.02 & 81.12 & 80.18 & 78.74 & 77.78 & 77.20 & 76.46 & 75.67 & 75.27 & 74.57 & 74.53\\

\bottomrule
\end{tabular}
}
\label{tab:detailed-CompCars}
\end{table*}

\section{{Computational Cost Analysis}}
\begin{table}[ht]
\centering

\caption{{Analysis of trainable parameter count and inference computational cost for different methods on the CUB200 dataset. The computational cost is measured with a batch size of 100.}}
\renewcommand{\arraystretch}{1.3}
\resizebox{\textwidth}{!}{
\label{comparison-table}
\begin{tabular}{lcccc}
\toprule
Method & Model Parameters (M) & Inference Cost (GFLOPs) & Center Parameters (M) & ANJ Cost (GFLOPs)\\
 \midrule
Decoupled-Cosine\cite{NIPS2016_vinyals} & 11.18 & 182.35   & +0.0512 & +1.68  \\
FACT\cite{zhou2022forward}    & 11.18 & 182.35   & +0.0512 & +1.68  \\
SAVC \cite{song2023learning}    & 11.50  & 364.77   & +0.1024 & +3.27 \\

\bottomrule
\end{tabular}
}
\label{tab:computational_cost}
\end{table}
{We analyze the computational cost of SAAN compared to other methods on CUB200 dataset. The additional cost introduced by SAAN mainly comes from the increased parameters of trainable centers and the inference computation brought by ANJ. Therefore, we report the extra training parameters and inference computation introduced by SAAN, as well as the parameter count and inference computation of other methods themselves, as shown in Tab. \ref{tab:computational_cost}. It can be observed that the additional parameter count and inference computation introduced by SAAN are both less than 1\%. In SAVC, the computational cost and center parameter count are both twice as high as in other methods because SAVC performs category augmentation with virtual classes, leading to a doubled number of centers. During inference, both the original and augmented samples need to be processed, resulting in increased inference computation.}

\section{Repeated Experiments on 2SNCM}
\label{sec:2SNCM}
\begin{table*}
\centering
\caption{Comparison 2SNCM with NCM on three mainstream datasets in FSCIL. $\Delta_{avg}$: The average of the accuracies on all sessions. The data comes from the average accuracy of 5 random experiments.}
\resizebox{\textwidth}{!}{%
\begin{tabular}{cccccccccccccc}
\toprule
\multirow{2}{*}{Dataset} & \multirow{2}{*}{Method} & \multicolumn{11}{c}{Acc. in each session (\%)$\uparrow$} & \multirow{2}{*}{$\Delta_{avg}$}\\
\cline{3-13} & & 0 & 1 & 2 & 3 & 4 & 5 & 6 & 7 & 8 & 9 & 10 \\
\hline
CIFAR100 &NCM & $74.17$ & $68.80$ & $64.33$ & $60.37$ & $57.06$ & $53.92$ & $51.71$ & $49.68$ & $47.66$  &--&--& 58.63\\
CIFAR100 &2SNCM & $74.35$ & $69.03$ & $64.56$ & $60.60$ & $57.30$ & $54.11$ & $51.81$ & $49.65$ & $47.70$  &--&--& 58.79\\
\hline
CUB200 &NCM&   73.18	&68.41 & 64.32 &60.05 &58.67 &54.71 &53.02 &50.42 &49.38	&48.09	&46.79 &57.00 \\
CUB200 &2SNCM&  73.28	&68.58&64.51&60.24&58.86&54.88&	53.19&50.58&49.52&48.17&46.93
 &57.16 \\
\hline
\emph{mini}ImageNet &NCM & 70.32 & 65.29 & 61.25 & 58.01 & 55.09 & 52.11 & 49.51 & 47.66 & 46.01  &--&--& 56.14\\
\emph{mini}ImageNet &2SNCM& 70.46&65.33&61.27&58.01&55.04&52.04&49.42&47.59&45.91
 &--&--& 56.12\\

\bottomrule
\end{tabular}
}
\label{tab:comparison_ncns}
\end{table*}
2SNCM and NCM only differ in the selection of representative points. 
In order to exclude random perturbations, we conducted multiple repeated experiments on them.
Here the seeds are canceled and multiple repeated experiments are conducted.
We use Decoupled-Cosine to conduct 5 replicate experiments on 3 datasets respectively and report the average performance of them in Tab. \ref{tab:comparison_ncns}.
Compared with NCM, 2SNCM improves the average accuracies on CIFAR100, CUB200 by 0.16\%.
On \emph{mini}ImageNet, 2SNCM performs better in the early incremental phase and base phase, and NCM performs better in the late incremental phase.
The better performance of 2SNCM indicates the effectiveness of normalizing based on angles when there are enough base round samples while preserving norm information without normalization when the incremental round samples are fewer.



 \bibliographystyle{elsarticle-num-names} 
 \bibliography{nn}






\end{document}